%% file: article.tex
\documentclass[3p]{elsarticle}
\usepackage[T2A]{fontenc}
\usepackage[utf8]{inputenc}

\usepackage{graphicx}
\usepackage{amsfonts}
\usepackage{color}
\usepackage{hyperref}
\usepackage[ocgcolorlinks]{ocgx2}
\usepackage{tabularx}
\usepackage{multirow}
\usepackage{caption}
\usepackage{subcaption}
\usepackage{amsthm}
\usepackage{amsmath}
\usepackage{amssymb}
\usepackage{mathtools}
\usepackage{xpatch}
\usepackage[nodisplayskipstretch]{setspace}

\usepackage{placeins}
\usepackage{float}

\begin{document}

\begin{frontmatter}
\input{front_matter.tex}
\end{frontmatter}

\makeatletter
\patchcmd{\ps@pprintTitle}{Preprint}{\footnotesize\itshape Accepted manuscript. Published in \textit{Neural Networks}. \hfill DOI: \href{https://doi.org/10.1016/j.neunet.2026.109502}{10.1016/j.neunet.2026.109502}}{}{}
\makeatother

\section{Introduction}
\label{sec:intro}

Deep neural networks are extensively used to solve a number of applied problems in various fields, such as computer vision~\citep{levi2015age, Wang_2015_ICCV, ren2016faster, falk2019u, Sadeghian_2017_ICCV}, signal and audio processing~\citep{bai2021speaker, ren2019fastspeech, akbari2021vatt, rim2020deep}, and natural language processing~\citep{cho2014learning, bahdanau2015neural, young2018recent}. Being nonlinear multi-parametric models, they demonstrate efficient performance but are often employed as a black box. Numerous efforts to shed light on
how these networks function have led to the establishment of an actively developing area of explainable artificial intelligence~\citep{10.3389/fdata.2021.688969, linardatos2021explainable}.

More importantly, artificial neural networks have already extended beyond the abstract biologically inspired models built to solve classical machine learning problems. They are actively used as a proxy for neural coding and decoding of their biological origins. For example, internal activations of deep neural networks have been successfully related to neural activity in the auditory pathway, revealing convergence between artificial model representations and neurophysiological activity~\citep{li2023dissecting}. The evolution of artificial models' capabilities means they can be studied by applying methods and paradigms established in neuroscience, particularly the treatment-control paradigm. Indeed, one can feed specific inputs or distort the inference and measure the resulting changes in activations. Employing a single treatment, though, is restrictive. Thus, considering a series of multiple treatments applied simultaneously better matches real-world conditions.

An intentional modification to the input is known as augmentation. Since image data transformations can be prominently traced and displayed, classic computer vision tasks, including image classification, appear to be a valuable tool for developing methods to study neural networks. In this context, augmentations can be regarded as actors modifying certain properties of the world as it is perceived by the network, such as changes in colours, rotations, mirror reflections, and others. And a real-world scenario for computer vision involves various environmental changes that are correlated and dependent on each other. To distinguish between outer (environmental) and inner (model-specific) interactions, one can estimate the effects with a baseline model as a reference point for further study.
This approach raises the trade-off modelling problem: selection should be capable of sensing external changes while remaining adequately simpler than the original model. At the same time, this problem can be mitigated by introducing additional variables accounting for possible interactions at the expence of increased dimensionality of the analysis.

As for complexity, modern deep learning systems contain several connected modules, often following the declarative paradigm in their architectural design, according to which the necessary functioning limits are specified (output dimensionality, number of blocks or layers, their hyperparameters, stabilizing modules, etc.), and high-level demands are imposed to guide a network (e.g., sequential redistribution of information from the spatial domain to separate channels). It appears to be important to study these internal modules and their activations during inference to identify the network's stability modes and operational states, similar to a learning stability analysis (regarding, e.g., vanishing gradients or gradient explosion~\citep{pascanu2013difficulty}). A well-known research area in this context is related to adversarial attacks, which disrupt the normal mode of  operation. In the long term, this line of research could make it possible to eliminate redundancy and develop more compact and efficient models, presumably due to their greater specialization. In practice, it is already established that trained networks are compressible via quantization~\citep{3600270.3602468} or through statistical methods like variational dropout~\citep{molchanov2017variational}.

A reasonable descriptor of a neural network's functioning is variance, both in terms of predictions and internal activations. Adjustments to the input data might be considered sources of variance; and as soon as multiple sources act simultaneously, it is natural to assess the contribution of each source~\citep{saltelli2008global}. Considering the variance as uncertainty measure leads to sensitivity analysis, where these contributions are assigned as sensitivity values.

\begin{figure}[t]
    \includegraphics[scale=0.55]{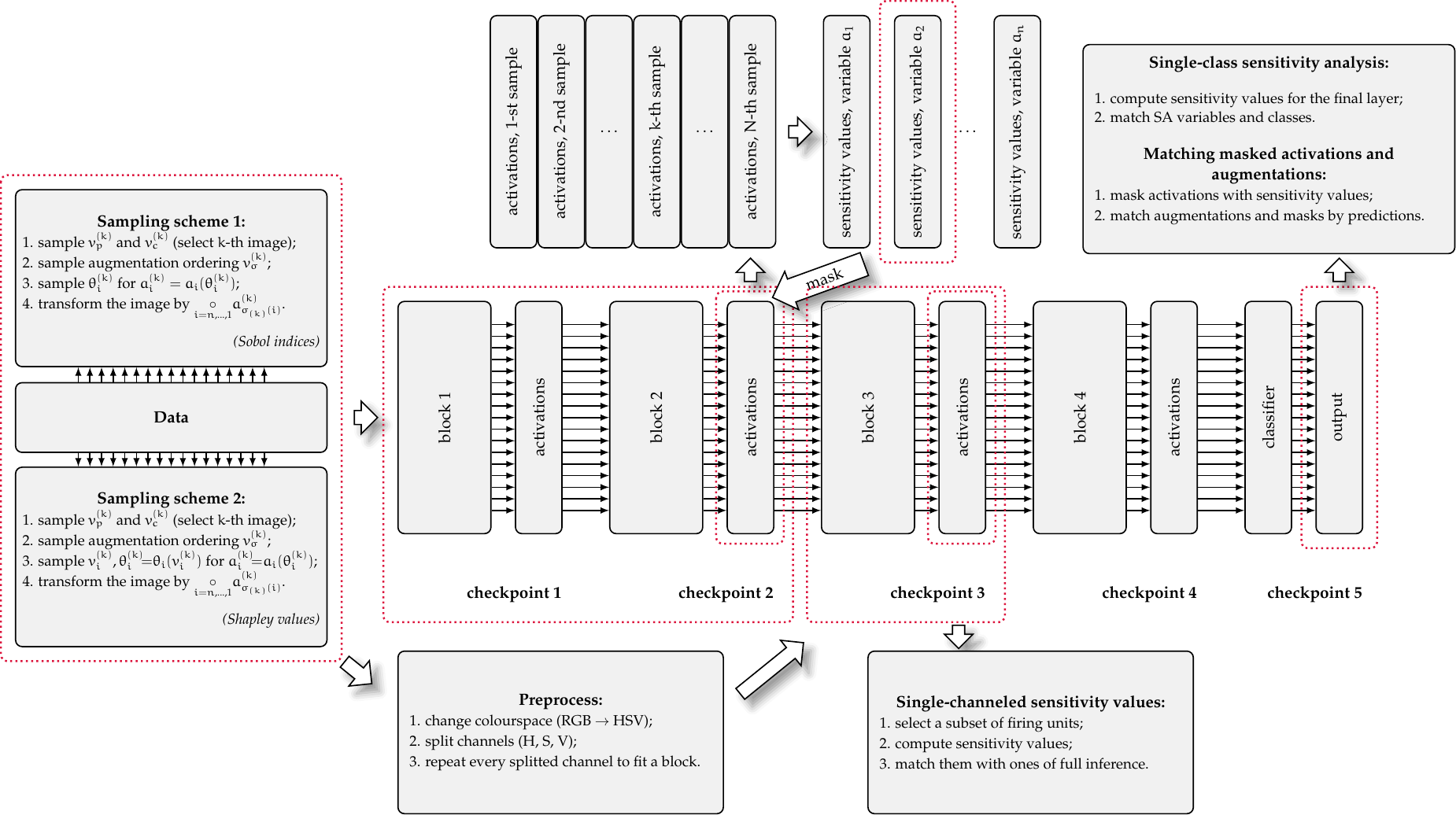}
    \caption{Summarized pipeline of the proposed framework. A block refers to a segment of a neural network participating in the analysis as a single mapping from its input to the corresponding activations. A checkpoint is a point of the network at which activations are gathered; as it is related to the preceding block, both terms are used interchangeably. }
    \label{fig:0-pipeline}
\end{figure}

The current work employs sensitivity analysis based on Sobol indices and Shapley values to highlight the network components that provide the most prominent feedback in response to input augmentations. As a result, a framework for selecting and studying units
sensitive to input data augmentations has been established (Figure~{\ref{fig:0-pipeline}}). The variables incorporated in the analysis are augmentation-related, augmentation-permuting, partition- and class-selecting ones.

Three major experimental setups have been developed: (1) computing and analyzing ``raw'' sensitivity values; (2) relating sensitivity variables to single augmentations via guided masking prediction; (3) exploring correlations between isolated segments of a network.

Within the series (1), raw sensitivity values are explored by visual inspection and further analyzed using statistical tools, treating them as features or variables, to identify their internal connections and trends from different perspectives. Linear discriminant analysis (LDA) and analysis of Spearman correlations based on sensitivity values provide complementary insights into separability and specificity of these values, as well as possible relationships among units aggregated within the spatial dimension. Spatial maps emerge as sensitivity values grouped by a sensitivity analysis variable in modules that maintain the spatial structure of their output. Depending on the augmentation, such images reveal spatial structures that may be associated with the specifics of the transformation (e.g., central vertical line for a randomly applied horizontal flip, formed by values that differ from the background). Separately computed spatial correlations reflect the relationships between sensitivity values and coefficients of variation as a  module performance medium.

Experimental series (2) utilizes pre-computed sensitivity values for masking corresponding activations of the checkpoints during inference. Independently, noise from augmentations applied solely is injected, and the resulting predictions generate a signal related to the combined influence of the masks and augmentations. This part of the work also addresses specificity but in a completely different way, by involving the inference process and seeking instabilities caused by the guided masking. In addition, a single-class sensitivity analysis has been derived as an effort to understand the sensitivity of the entire network on the basis of the classification layer. The sensitivity values of its outputs possibly reflect instabilities in predictions related to input augmentations. As evidence that these findings are generalizable, the constructed pipeline seeks prediction bias as changes in the frequency of predictions. These changes are related to classes sensitive to an augmentation as observed at the classification layer, and the prediction bias is generated by the guided masking of convolutional checkpoints.

In the last experimental series, sensitivity values are used to evaluate internal connections between network segments operating independently from each other. The initial purpose of this analysis is to examine how well patterns extracted for isolated segments can be matched with the same patterns for a normally operating network. Network segments are related to each other through correlations of their sensitivity values, revealing higher-level patterns that highlight the similarities and differences among them.

Experiments on AlexNet~\citep{krizhevsky2012imagenet}, VGG11~\citep{simonyan2015}, and ResNet18~\citep{he2016deep} models, supplemented by additional findings from vision transformers ViT-b-16~\citep{ViT_ICLR2021dosovitskiy} and Swin-t~\citep{LiuL00W0LG21}, have been conducted in order to demonstrate alignment among different analysis blocks and consistency with external knowledge, including patterns observed in biological circuits. The authors also recognize these findings as potentially transferable to the study of biological neural networks in complex environments.

\section{Related works}

As mentioned in the~\hyperref[sec:intro]{Introduction}, possible applications include research on the robustness of deep-learning model inference. In~\citep{hendrycks2019robustness} the authors introduced a comprehensive benchmark for analyzing the stability of predictions using perturbed ImageNet images and identified connections between adversarial and unconditioned augmentations similar to those considered in the current work (e.g., pixelation, Gaussian blur, impulse noise, snow, and others). Their approach lacked interactions among different augmentations, as they were applied in an isolate manner.
In their subsequent work~\citep{hendrycks2021nae} the authors approached the problem from a different angle and proposed a dataset containing natural adversarial examples that are constantly misclassified by many deep learning models. Disentangling such images by means of simple augmentations may lead to new architectural designs or enhance learning procedures, and in the context of the proposed framework, it can be approached by analyzing the influence of simultaneously applied augmentations.

In a broader perspective, many real-time applications feature heterogeneous actors exerting various effects on the environment and operating agents. For all such examples, it is critical not only to robustly process the data but also operate under real-time conditions. That implies the necessity to maintain the continuity of the internal data flow within a system, in particular, to be rejection-tolerant. Such a property could be achieved in different ways, and by employing the proposed framework, one can implement quick corrections of the inference based on the estimated sensitivity values (guided masking). The same logic could be applied if one considers the separate dataflows as sources of variance themselves while integrating multiple modalities. For example, the authors of~\citep{goyal2025multi} built an ensemble framework by integrating multiple telemetric modalities, which could be treated as simultaneously operating secondary sources of variance prone to vulnerability to distribution shifts.
Such shifts could be identified and potentially neglected by utilizing pre-computed sensitivity maps of the corresponding variables.

The demand for assessing multiple augmentations is also demonstrated in~\citep{valois2024occlusion}, which introduces occlusion sensitivity analysis with a deep feature augmentation subspace. The proposed method enables integration of various augmentations into an occlusion sensitivity pipeline by constructing the reference basis as an orthonormalized set of outputs from the layer preceding the classification unit. The occlusion framework is then used to generate explanation heatmaps based on angles between the constructed subspaces. In other words, the proposed framework estimates pixel-wise significance heatmaps, potentially highlighting regions that are important for prediction. Augmentations were used as an enhancer to support the occlusion estimator. In the current study, augmentations are employed as an analysis medium, the influence of which is directly examined by simultaneously applying them to the input to produce sensitivity maps, not occlusion heatmaps. Moreover, most uses of the occlusion framework are designed for completely different purposes, as shown below.

Among the methods developed for explainable AI
are attribution methods, which identify the raw data components that most influence the network's decision-making, and methods for constructing interpretable
images that most activate its neurons. The latter include
Activation Maximization~\citep{erhan2009visualizing},
Deconvolutional Neural Networks~\citep{zeiler2014visualizing},
Network Inversion~\citep{mahendran2015understanding,dosovitskiy2016inverting},
and Network Dissection~\citep{bau2017network}, the last of which may also be
classified as an attribution method. 
Activation Maximization and visualization by Deconvolutional Neural Networks are designed to construct images to which distinct neurons react most strongly. Network Inversion recovers input images from their feature maps at different layers, making it possible to understand which information from the original feature space is retained in each layer. The Network Dissection method matches neurons with individual semantic parts of the original images. The capabilities of the methods have been demonstrated in tasks such as representation analysis in neural style transfer and visualization of adversarial noise and adversarial examples~\citep{qin2018convolutional}.

An important group of methods for studying data processing models is sensitivity analysis, which seeks to estimate the contribution of the selected variables' degree of uncertainty to the total uncertainty of the mapping~\citep{saltelli2008global}. On the whole, sensitivity analysis belongs to a broader class of variable importance methods (VIMs). The corresponding variable importance analysis~\citep{wei2015variable} includes difference-based VIMs, parametric and non-parametric regression techniques, hypothesis testing techniques, variance-based and moment-independent VIMs, and graphical VIMs. Special attention is to be paid to the variance-based VIMs as variance naturally incorporates both the firing specifics of the network's inference and the impact of input modifications (treatments). More specifically, variance-based VIMs such as Sobol indices~\citep{sobol2001global} and Shapley values~\citep{song2016shapley} decompose the variance into components related to subsets of variables controlling the input augmentation.

Both sensitivity analysis methods and various visualization techniques are typically considered for low-level features, e.g., distinct pixels. In this work, variance-based VIMs measuring the influence of high-level variables were considered to significantly reduce the dimensionality of the feature space. Moreover, many artificial intelligence models do not directly relate to high-level concepts familiar or understandable to humans~\citep{kim2018interpretability}, which significantly complicates analysis by means of low-level variables. As the study focuses on explicit computational models, considering a limited number of fully controlled high-level variables in order to explore and hypothesize the possible connections between internal activations and treatments, the proposed approach is minimally affected by the drawbacks of Sobol indices~\citep{wei2015variable}.

However, computing the Sobol indices in the case of dependent variables is complicated due to the necessity to account for correlated parts. The Shapley values-based approach addresses the issue automatically but has its own implementation difficulties. \cite{lundberg2017unified} proposed options for the practical use of Shapley values as a method for sensitivity analysis of computational models, including a special case for linear models (Linear SHAP), methods with simplified computation of expectations (Low-order SHAP, Max SHAP), and the DeepSHAP method that combines DeepLIFT approach~\citep{shrikumar2017learning} and Shapley values. In addition, \cite{lundberg2017unified} have constructed a model-independent KernelSHAP method based on Linear LIME~\citep{ribeiro2016should}, i.e., weighted least-squares between computational and linear explanation models with Shapley values used as weights.

Compared to certain existing explainability methods, the current work proposes to focus on how the network propagates the input distortions themselves instead of using them to build attribution maps for specific image samples (e.g.,~\cite{zeiler2014visualizing,ribeiro2016should,petsiuk2018rise,fel20look21}). Although there are studies that also explore the influence of various distortion types~\citep{samek2016evaluating,geirhos2018generalisation,hendrycks2019robustness,vilone2021notions,bachoc2023explaining}, to the best of the authors' knowledge, simultaneous input augmentations have not yet been presented. The authors of the current work see their main contribution as addressing this issue and proposing a framework that is able to account for potential dependencies among injected distortions, as well as other variables affecting the input data, such as class or data partition. As dependencies arise even from non-permutability of the sequential augmentations, their order is another crucial factor incorporated into the analysis.

\section{Materials and methods}
\label{sec:mat-met}

\subsection{Convolutional neural networks}

By artificial neural network, one usually means a parametric mapping with a layered structure, every single layer of which is a linear transformation of inputs followed by an element-wise nonlinearity. As an example, a feedforward network may be mentioned, $f(x) = l_N[l_{N-1}[\ldots l_1[x]]]$, where $l_i[z] = g_i(W_i z + b_i)$ are the layers, $W_i$ are matrices of complementary sizes, $b_i$ are bias vectors, $g_i(\cdot)$ are element-wise nonlinearities, and $i=\overline{1, N}$. A layer is composed of several individual ``neurons'', mathematical models of biological neurons, which are parameterized by single rows of the matrix $W$ and the corresponding components of the bias vector $b$.

Convolution networks are artificial neural networks in which the linear part of the majority of layers is parameterized by convolutions, $W[x]: \mathbb{R}^{c_{\text{in}} \times n_1 \times \ldots \times n_d} \to \mathbb{R}^{c_{\text{out}} \times m_1 \times \ldots \times m_d}$,
\begin{equation}
\label{eq:cnn-conv-operation}
\begin{aligned}
y_{i, s_1, \ldots, s_d} = \sum\limits_{j} w_{i, j, \cdot} \ast x_{j, \cdot} = \sum\limits_{j} \sum\limits_{1 \leq k_1 \leq n_1} \ldots \sum\limits_{1 \leq k_d \leq n_d} w_{i, j, k_1, \ldots, k_d} x_{j, s_1 - k_1, \ldots, s_d - k_d}, \\
w \in \mathbb{R}^{c_{\text{out}} \times c_{\text{in}} \times k_1 \times \ldots \times k_d}, \, s_t = \overline{1, m_t}, \, t = \overline{1, d}.\\
\end{aligned}
\end{equation}

In the equation above, the stride and dilation parameters were omitted for simplicity (by default, their values are 1 and 0 respectively). The stride parameter determines the offset between neighbouring blocks passed to the convolution receptive field, and the dilation controls the offsets among elements to be grouped as a single block.
Non-strided and non-dilated layers of this type are shift-equivariant, meaning they can react to the target subvector (submatrix, subtensor) regardless of its position, with the corresponding activation being only shifted. This property is similar to the visual system's ability to recognize sensed objects regardless of binding their position to a specific point in the visual field.

Among other parallels that could be drawn between convolutional layers and the human visual system, it is worth noting the following points: (1) the local connectivity of convolutional layers and the local connections between cells in the visual pathway (retina, lateral ganglion nucleus, striate cortex V1)~\citep{purves2018neuroscience,kandel2021principles}, which determine the presence of receptive fields; (2) the visual cortex organization, where areas beyond the primary visual cortex process higher-order visual information, while the primary visual cortex (V1) itself reacts to the low-level features such as the orientation or direction of local edges~\citep{purves2018neuroscience,kandel2021principles}; and the organization of CNN layers, where the deeper the layer, the more complex the stimuli required to elicit high activation~\citep{zeiler2014visualizing,lindsay2021convolutional}. From a computational point of view, convolutional layers are also attractive because they have fewer parameters than fully-connected ones and thus require fewer computational resources for learning and inference.

For classification tasks, where an artificial neural network is to be trained on some training set to recognize labels of input samples, variations of the following architecture are common. It begins as a stack of several convolutional layers, with the number of filters increasing with depth, followed by several fully-connected layers, the last of which has the number of outputs equal to the number of classes. Such an architectural design forces the output to gradually decrease in spatial dimensionality while increasing the number of channels. Actual deep learning models also contain various submodules
to improve the computational stability of learning (e.g., batch normalization), prevent overfitting (dropouts), and streamline other aspects of the data flow.

Known architectures of such networks include AlexNet~\citep{krizhevsky2012imagenet}, VGG~\citep{simonyan2015}, and ResNet~\citep{he2016deep}. A feature of the last-mentioned family is the presence of skip connections, allowing such networks to reuse data representations from previous layers within deeper structures. This property makes the ResNet-like networks more similar to biological neural networks (for example, in the visual cortex, there are V1 $\to$ V4 and V1 $\to$ MT shortcuts in addition to V1 $\to$ V2 $\to$ V4 and V1 $\to$ V2 $\to$ MT paths~\citep{kandel2021principles}).

\subsection{Sobol indices}
\label{subsect:sobol-indices}

Sobol indices are closely related to ANOVA and serve as values that describe the amount of variance explained by different subsets of variables~\citep{sobol2001global, gasanov2020sensitivity}. Sensitivity analysis of this type requires a particular real-valued function
that depends on several variables.
In neural networks, a single neuron can constitute such a mapping.

Let $x_1, \ldots, x_d$ be a set of independent variables (factors), for which one considers a function $f: \mathbb{R}^{d} \to \mathbb{R}$.
To avoid cluttering formulas with indices, let us introduce the following designation,
$\alpha = \{ j_1, \ldots, j_{|\alpha|} \} \subset D \equiv \{1, \ldots, d\}$, where $|\alpha|$ is a number of elements in the set, and one then can denote a subset of variables with respect to $\alpha$ as
$x_{\alpha} = (x_{j_1}, \ldots, x_{j_{|\alpha|}})$.
Let us also write the complement of $\alpha$ to the set $D$ as
$\overline{\alpha} = D \setminus \alpha$.

Taking into account Sobol's derivation and Hoeffding's decomposition~\citep{hoeffding1948class} with the conditions
$E_{x_{\beta}}[f_{\alpha}(x_{\alpha})] = 0$ $\forall \beta \subset \alpha \subseteq D$,
one can arrive at the following unique representation of such a function:
\begin{equation}
\label{eq:sobol-decomposition}
f(x_1, \ldots, x_d) = \underbrace{\vphantom{f_{i j}(}f_0}_{\text{constant}} + \sum\limits_{1 \leq i \leq d} \underbrace{\vphantom{f_{i j}(}f_i(x_i)}_{\text{major effect}} + \sum\limits_{1 \leq i < j \leq d } \underbrace{f_{i j}(x_i, x_j)}_{\text{2nd-order effect}} + \ldots = \sum\limits_{0 \leq s \leq d} \sum\limits_{\alpha: |\alpha|=s} f_{\alpha}(x_{\alpha}),
\end{equation}

\noindent where the internal summands can be written as
\begin{equation}
\label{eq:sobol-decomposition-terms}
\begin{aligned}
f_0 = E[f(x)], \quad f_i(x_i) = E_{x_{\overline{\{i\}}}}[f(x)|x_i] - f_0, \\
f_{i j}(x_{i}, x_{j}) = E_{x_{\overline{\{i,j\}}}}[f(x)|x_i, x_j] - (f_0 + f_i(x_i) + f_j(x_j)), \quad \ldots; \\
\end{aligned}
\end{equation}

\noindent and in the general case, they can be found from the equations
$f_{\alpha}(x_{\alpha}) = E_{x_{\overline{\alpha}}} [ f(x) | x_{\alpha}] - \sum\limits_{\beta \subsetneq \alpha} f_{\beta}(x_{\beta})$.
Moreover, using this decomposition, one can construct the following representation of the mapping's variance:
\begin{equation}
\label{eq:sobol-variance-decomposition}
V[f(x)] = E\left[\left( f(x) - E[f(x)] \right)^2\right] =  \sum\limits_{1 \leq i \leq d} V_{i} (f(x)) + \sum\limits_{1 \leq i \leq d} \sum\limits_{1 \leq i < j \leq d} V_{i j} (f(x)) + \ldots = \sum\limits_{\alpha \subseteq D} V_{\alpha} (f(x)),
\end{equation}

\noindent with the partial variances
$V_{\alpha} (f(x)) =
V_{x_{\overline{\alpha}}} \left[ f_{\alpha}(x_{\alpha}) \right]$.
Hereby, the Sobol indices are defined as
\begin{equation}
\label{eq:sobol-variance-decomposition-term}
\begin{aligned}
S_{\alpha}(f(x)) = \frac{\sum\nolimits_{\beta \subseteq \alpha} V_{\beta} (f(x))}{V[f(x)]} = \frac{ V_{x_{\alpha}} \left[ E_{x_{\overline{\alpha}}}[f(x) | x_{\alpha}] \right]}{V[f(x)]}, \\
S_{\alpha}^{\text{T}}(f(x)) = \frac{\sum\nolimits_{|\alpha\cap\beta| > 0} V_{\beta} (f(x))}{V[f(x)]}
= 1 - \frac{V_{x_{\overline{\alpha}}}\left[ E_{x_{\alpha}}[f(x) | x_{\overline{\alpha}}] \right]}{V[f(x)]}, \\
\end{aligned}
\end{equation}

\noindent where $S_{\alpha}(f(x))$ is an ordinary Sobol index for a combination of variables $\alpha$, and $S_{\alpha}^{\text{T}}(f(x))$ is the so-called total Sobol index for the same combination. The difference between them is as follows: ordinary Sobol indices indicate the fraction of variance explained by a single combination, whereas the total indices accumulate the variance fractions of all combinations that contain $x_{\alpha}$. Total effect indices are used in many-variable cases to reduce the computational complexity. There are also the following relations between these indices:
$0 \leq S_{\alpha}(f(x)) \leq S_{\alpha}^{\text{T}}(f(x)) \leq 1$.

Hereinafter, the ordinary Sobol indices for a single variable, $\alpha = \{j\}$, $j=\overline{1, d}$, are referred to as first-order Sobol indices, and, similarly, total Sobol indices refer to those for a single variable.

In practice, Sobol indices are computed by sampling within a mapping domain. This work relies on quasi-Monte Carlo methods, Sobol sequencing~\citep{sobol2001global} and Saltelli sampling~\citep{saltelli2010variance}, implemented in the SAlib Python package~\citep{Herman2017}.

Finally, multiple variables can be grouped~\citep{saltelli2002making}. For example, if one considers the augmentation parameters as independent variables, they can be directly aggregated by the corresponding augmentation. Such grouping substantially facilitates further analysis, allowing the exploration of interactions among higher-level instances.

\subsection{Shapley values}

However, the statistical independence of the variables may be violated in practice, making the ANOVA variance decomposition (\ref{eq:sobol-variance-decomposition}) invalid. In this case, its generalization (ANCOVA) is to be considered, with corresponding changes in the Sobol indices~\citep{chastaing2015generalized}:
\begin{equation}
\label{eq:sobol-ancova-decomposition}
V[f(x)] =  \sum\limits_{\alpha \subseteq D} \underbrace{\vphantom{\left(\sum\nolimits_{\beta}\right)}\text{Cov} \left( f(x), f_{\alpha}(x_{\alpha}) \right)}_{=V[f(x)] \cdot S_{\alpha}} = \sum\limits_{\alpha \subseteq D} \Big( \underbrace{\vphantom{\left(\sum\nolimits_{\beta}\right)}V_{\alpha} [f(x)]}_{=V[f(x)] \cdot S_{\alpha}^{u}} + \underbrace{\text{Cov} \left( f_{\alpha}(x_{\alpha}), \sum\nolimits_{\beta \subseteq \overline{\alpha}
} f_{\beta}(x_{\beta}) \right)}_{=V[f(x)] \cdot S_{\alpha}^{c}} \Big),
\end{equation}

\noindent where $\text{Cov}(z_1, z_2)$ is the covariance between $z_1$ and $z_2$, $S_{\alpha}$ is the analogue of the standard Sobol indices, which are decomposed into the sum of the variances from uncorrelated, $S_{\alpha}^{u}$, and correlated, $S_{\alpha}^{c}$, contributions. There are several difficulties in constructing this type of decomposition~\citep{chastaing2015generalized,li2010global}. Another alternative is to reconsider the basic computational model as a fully probabilistic one, expressed as a Bayesian network~\citep{ballester2022computing}; however, this requires providing a statistical model of the input data, which is a separate challenging task.

The authors of~\citep{song2016shapley} have proposed an approach based on Shapley values, which they refer to as Shapley effects. The interpretation of Shapley values is that they account for the contribution of each variable to the prediction, averaged over all possible combinations of the variables. Assume, as before, that there is a set of variables $\{x_i\}_{i=1}^{d}$ for which a cost (value) function $c: 2^d \to \mathbb{R}$, $c(\varnothing) = 0$, is considered, determining the significance of the contribution of each variable combination. Under these assumptions, the Shapley value for a single variable is defined as
\begin{equation}
\label{eq:shapley-value-def1}
    v_i = \sum\nolimits_{\alpha \subseteq D \setminus \{ i \}} w_{\alpha} \left( c(\alpha \cup \{i\}) - c(\alpha) \right), \quad w_{\alpha} = \frac{(d - |\alpha| - 1)! |\alpha|!}{d!} = \frac{1}{d} \left[ C_{d-1}^{|\alpha|} \right]^{-1}.
\end{equation}

The cost function should be chosen to relate the variables to their contribution to the total variance. These can be the numerators of the Sobol indices:
\begin{equation}
    c_1(\alpha) = V_{x_{\alpha}} \left[ E_{x_{\overline{\alpha}}}[f(x) | x_{\alpha}] \right], \quad 
    c_T(\alpha) = V[f(x)] - V_{x_{\overline{\alpha}}}\left[ E_{x_{\alpha}}[f(x) | x_{\overline{\alpha}}] \right] = E_{x_{\overline{\alpha}}}\left[ V_{x_{\alpha}}[f(x) | x_{\overline{\alpha}}] \right],
\end{equation}

\noindent and the study~\citep{song2016shapley} shows that the selection of these cost functions results in equivalent Shapley values; as the authors note, $c_T(\alpha)$ is preferable when used within the Monte Carlo estimator.
An essential feature of Shapley values derived in this way is that they always sum to the total variance.

For practical computation, the formula (\ref{eq:shapley-value-def1}) can be rewritten in terms of permutations as
\begin{equation}
    v_i = \sum\nolimits_{\pi \in \Pi(D)} \frac{1}{d!} \left( c(P_i(\pi)\cup\{i\}) - c(P_i(\pi)) \right),
\end{equation}

\noindent where $\Pi(D)$ is the set of all $d!$ permutations of variables from $D$,
$P_i(\pi) = \{\pi(1), \ldots, \pi(j) | \pi(j+1) = i\}$ are the variables in the permutation $\pi$ preceding the $i$th one. To calculate this value, the authors~\citep{song2016shapley} approximated it by sampling a subset of possible permutations and then averaging.

\subsection{Data augmentation}

Data augmentation refers to various transformations made to increase the training samples' size and population. Moderate augmentation allows the neural network to ``observe'' more diverse examples, acting as a regularizer to reduce overfitting. Augmentation is used both for small-sample problems and for improving learning stability on large datasets, and has also been related to dreaming~\citep{shorten2019survey}, one of the essential human mental functions.

According to the authors~\citep{shorten2019survey}, existing augmentation methods can be roughly divided into warping (transforming existing samples) and oversampling (generating new ones). In the imaging domain, they have mentioned the following examples of warping: various geometric transformations that reduce positional/translational bias; colour space (photometric) transformations that minimize lighting bias; various types of blurring that enhance resistance to motion blur; cropping that addresses scaling variability; and random erasing that improves resistance to occlusion and intensifies exploration of the feature space. Oversampling methods operate on data distribution and include the following: k-Nearest Neighbours (k-NN) interpolation; guided data learning with generative adversarial networks (GANs); and neural style transfer, which switches conceptual data parameters such as day/night, winter/summer, artist style, and others.

The application of particular augmentation methods requires certain critical conditions to be considered. First, there must be some prior knowledge about the nature of the data, with which new samples must comply. For example, the typical camera orientation makes it meaningless to flip images vertically. Limiting the fraction of new samples is no less important. In particular, excessive oversampling on small datasets suppresses true variability in the data, since it is difficult to account for all possible sources of variance in this case. Augmentation should not significantly skew the data distribution or notably distort it in any other way; otherwise, the overfitting issue will resurface.

\subsubsection{Sampling schemes and variables for sensitivity analysis.}

Data augmentation causes changes in the pattern of neuronal activations. Similar to that in biological neurons, a specialization may develop among the units of a network, manifesting through these changes. Such specialization can be considered at the level of single neurons (as in the inferior temporal cortex, where individual neurons are likely involved in face, hand, or facial expression recognition~\citep{kandel2021principles}, with the clarification that these neurons are components of the network and receive preprocessed input), as well as at the level of neuronal groups (e.g., increased regional cerebral blood flow in V4 during the representation of abstract coloured figures, or V5/MT activation in response to visual motion patterns~\citep{kandel2021principles,zeki1991direct}). This work proposes using augmentations within sensitivity analysis to test these assumptions for deep learning models, starting with the warping transformations, which will hereinafter be referred to simply as augmentations.

Each transformation depends on its own set of parameters $\boldsymbol{\theta}_{i}$, $i=\overline{1, d}$, which are treated as independent variables. Augmentations without internal parameters (such as horizontal flip) should be supplied with a switching parameter that determines whether to apply the transformation.

Augmentations are applied to images sampled from a dataset divided into classes, and a classifier recognizes the corresponding labels. Therefore, the sensitivity analysis model is extended with a categorical class variable $v_c \sim \text{Cat}(1, \ldots, C)$ that selects the class from which an input is sampled. Considering separate variables for each class would significantly increase the dimensionality of the input space, making the approach impractical.

In supervised learning, the dataset is divided into non-overlapping training and validation subsets, which are used for fitting model weights and for tuning network hyperparameters and conducting various tests, respectively. The corresponding auxiliary variable was added to the analysis, $v_p \sim \text{Cat} \left( \{ \text{train}, \text{valid} \} \right)$, $p(v_p = \text{train}) = p(v_p = \text{valid}) = 1/2$, to capture possible differences in sensitivity depending on whether a sample was previously seen or unseen by the model.

A key point regarding augmentations is that the results of successive data transformations depend on their order. As an example, let us consider the following pair: $g_1(\cdot | \alpha)$ rotates the image by an angle $\alpha$, and $g_2(\cdot | \boldsymbol{\theta}_{g_2})$ erases a rectangular region---obviously, $g_1(g_2(\cdot | \boldsymbol{\theta}_{g_2}) | \alpha) \neq g_2(g_1(\cdot | \alpha) | \boldsymbol{\theta}_{g_2})$. Therefore, to account for the order of augmentations, the framework model has been enhanced with an additional variable $v_{\sigma} \sim \text{Cat}(\Pi(D))$, $p(v_{\sigma} = \widehat{v_{\sigma}}) = 1/d!$, whose realization, with equal probability, determines one of the possible permutations of the $d$ transformations. If certain combinations of augmentations are order-invariant, the number of possible permutations is reduced.

The resulting probabilistic model of sensitivity analysis variables is
\begin{equation}
\label{eq:input-space-model-1}
    p(v_c, v_p, v_{\sigma}, \boldsymbol{\theta}_1, \ldots, \boldsymbol{\theta}_d) = p(v_c) p(v_p) p(v_{\sigma}) \prod\nolimits_{i=1}^{d} p(\boldsymbol{\theta}_i), \\
\end{equation}

\noindent - in this case, all variables are independent, justifying the use of the ANOVA setting. Since each transformation modifies the data, the total number of augmentations should be limited. Excessively augmented samples will likely lose their intrinsic properties, resulting in a loss of representativeness of the original data.

To address this issue, switching variables, $\{\gamma_i\}_{i=1}^{d}$, which indicate whether a transformation is applied, can also be introduced to supplement the probabilistic model of the input space.
If a transformation $g_i(x)$ is to be ignored, $\gamma_i = 0$, this can be considered as applying the identity transform, or equivalently, selecting parameters $\boldsymbol{\theta}_{i, 0}$ that turn it into the identity, $g_i(\cdot | \boldsymbol{\theta}_i = \boldsymbol{\theta}_{i, 0}) \equiv \text{id}$.

The resulting probabilistic model is of the spike-and-slab type~\citep{ishwaran2005spike}:
\begin{equation}
\label{eq:input-space-model-2}
\begin{aligned}
p(v_c, v_p, v_{\sigma}, v_1, \boldsymbol{\theta}_1, \ldots, v_d, \boldsymbol{\theta}_d) = p(v_c) p(v_p) p(v_{\sigma}) \prod\nolimits_{i=1}^{d} p(v_i, \boldsymbol{\theta}_i), \\
p(v_i, \boldsymbol{\theta}_i) = (1 - \gamma_i) p_0(\boldsymbol{\theta}_i) + \gamma_i p(\boldsymbol{\theta}_i), \quad p_0(\boldsymbol{\theta}_i) = \delta(\boldsymbol{\theta} - \boldsymbol{\theta}_{i, 0}), \quad
\gamma_i = \mathbb{I}_{v_i > \tau_i}, \quad v_i \sim U[0, 1],\\
\end{aligned}
\end{equation}

\noindent where $\gamma_i$ is parameterized using uniformly distributed $v_i$, $0 < \tau_i < 1$ is a threshold specifying the probability of ignoring the augmentation, $\boldsymbol{\theta}_i$ are the transformation parameters, and $\delta(\boldsymbol{\theta} - \boldsymbol{\theta}_{i, 0})$ is the delta (point mass) distribution, $i=\overline{1, d}$.

In this formulation, independence between the variables is lost: the parameters of each augmentation depend on the switching variable. This fact makes the derived model (\ref{eq:input-space-model-2}) incompatible with ANOVA and Sobol indices. Shapley values were used for this setup since they naturally account for the case of dependent variables. In addition, a rigorous method for eliminating variable dependencies is described in~\cite{mara2015non}; however, it is based on the Rosenblatt transformation, which requires knowledge of the conditional cumulative distributions.

\subsubsection{Augmentation sets}

\begin{figure}[t]
    \hspace{-1 cm}
    \includegraphics[scale=0.4]{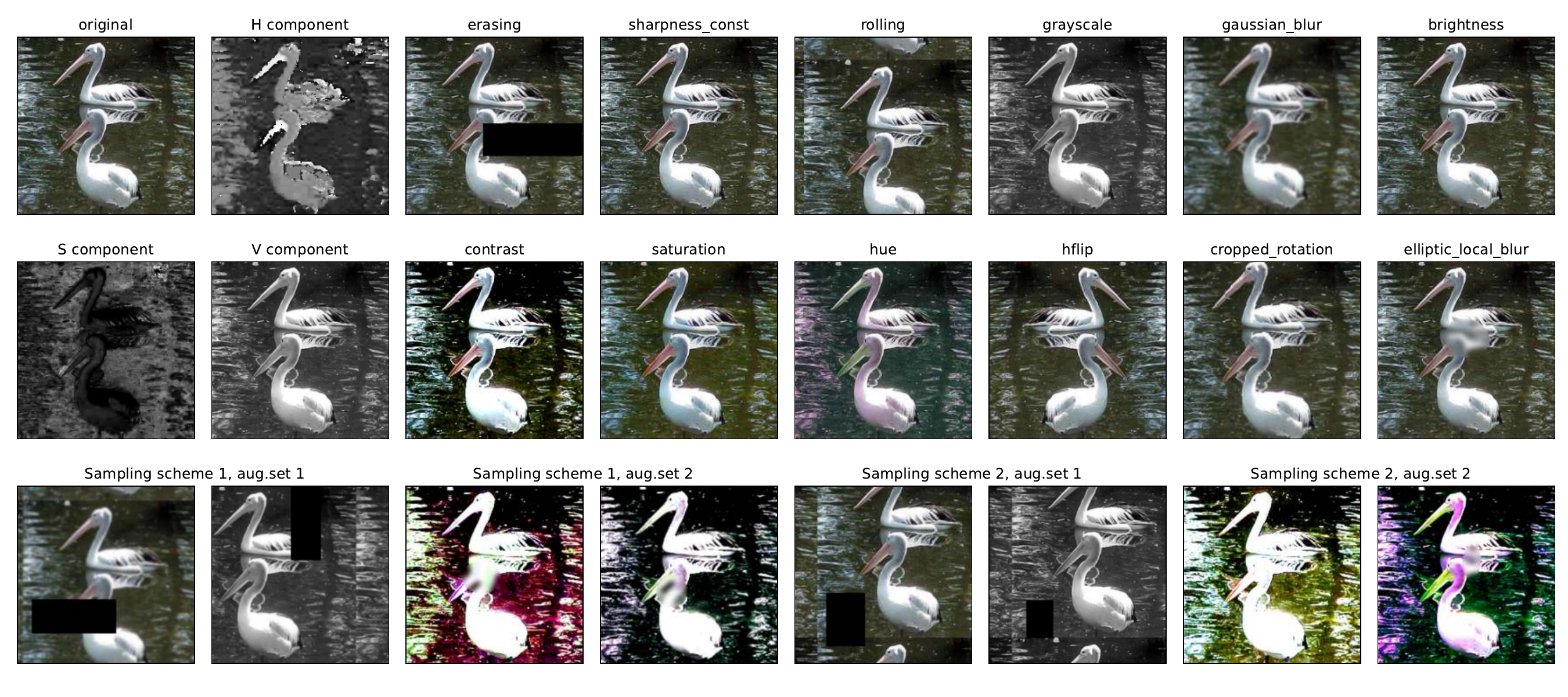}
    \caption{A sample image from the ILSVRC dataset, its corresponding HSV decomposition, and examples of single and simultaneous augmentations considered in the study.}
    \label{fig:1-augmentation_examples}
\end{figure}

Two separate sets of augmentations are considered, $\text{A}_1$ and $\text{A}_2$, (Figure~{\ref{fig:1-augmentation_examples}}). $\text{A}_0$ denotes the non-augmented (original) data.

The $\text{A}_1$ set includes five transformations:
($\text{A}_1.1$) erasing a rectangular part of an image,
($\text{A}_1.2$) adjusting sharpness by a fixed factor,
($\text{A}_1.3$) 2D rolling of pixels,
($\text{A}_1.4$) converting to grayscale,
and ($\text{A}_1.5$) Gaussian blurring.

The $\text{A}_2$ set consists of adjusting the following colour features:
($\text{A}_2.1$) brightness,
($\text{A}_2.2$) contrast,
($\text{A}_2.3$) saturation,
and ($\text{A}_2.4$) hue,
as well as applying
($\text{A}_2.5$) horizontal flipping,
($\text{A}_2.6$) rotation with successive cropping,
and ($\text{A}_2.7$) blurring of a random elliptic region in the centre of an image.

\FloatBarrier

\subsubsection{Sensitivity analysis (SA) variables}

In this study, the variable parameters $\boldsymbol{\theta}_i$ (or $v_i, \boldsymbol{\theta}_i$ in the~(\ref{eq:input-space-model-2}) case) of each augmentation were combined and treated as a single sensitivity analysis (SA) variable. The details of their configurations are listed in Table~{\ref{table:A0-details-aug-setup}}, including distributions and constant parameters. The auxiliary variables $v_c, v_p, v_{\sigma}$ were also considered as SA variables.

\subsection{Experimental setup}

The computations are divided into three experimental series: (1) computing sensitivity values for network components by using augmented data as input to the first layer; (2) damaging activations using the sensitivity values; (3) calculating sensitivity values for selected neurons in standalone segments using single channels of the original data represented in the HSV colour space. For the first and third experimental series, the number of activations and samples are described in~{\ref{appendix:exp-setup}}. Details related to the second series are provided below (Section~\ref{subsubsec:masked-actan}).

\subsubsection{Full-scale sensitivity analysis}

The first experimental series is designed as follows.
For each network, a set of submodules was selected to observe their activations (outputs). Hereinafter, they are called checkpoints and referred to as convolutional, transformer, or fully-connected if the last mapping before measuring their activations is a convolution block, transformer block, or fully-connected layer, respectively. During the experiment, data augmented by $A_k$, $k \in \{1, 2\}$ are fed into the input layer. The resulting activations measured at the selected checkpoints are saved for subsequent calculation of sensitivity values.

Sobol indices are computed employing the first feature space model (\ref{eq:input-space-model-1}), i.e., all augmentations are sequentially applied according to their sampled  permutation.
Internal parameters of the SA variables (for example, rotation angle, size and centre coordinates for rectangular erasing, etc.) are considered to be uniformly distributed over specified intervals and are sampled using the semi-random Sobol-Salteli method implemented in SAlib~\citep{Herman2017}.

The second input space model (\ref{eq:input-space-model-2}), designed for Shapley values, includes nominal variables. To sample from this distribution, as in the work~\citep{coqueret2017approximate}, independent samples of the nominal variables are drawn, indicating whether to apply a transformation or not. As all the remaining variables have no dependence structure, one may
choose either to use their default values (no augmentation) or to sample them from the corresponding marginal distributions.

Computed sensitivity values (Figures~{S2}--{S3}) were visually inspected and analyzed by extracting correlation and prediction patterns. The correlation patterns were evaluated by calculating unit-wise Spearman correlations between SA variables (Figures~{S1.2.1}--{S1.2.3}). Both augmentation sets were simultaneously considered in order to relate them. Units (output channels) of the fixed checkpoint were treated as observations. In the case of convolutional or transformer-block checkpoints, the results were spatially averaged. In addition to the resulting correlation matrices, spatial correlation maps were generated to reflect the possible relationships between sensitivity to augmentations and the coefficients of variation, used as a measure of unit activity (Figures~{S1.3.1}--{S1.3.3}).

Additional exploratory analysis was performed by extracting prediction patterns from confusion matrices (Figures~{S1.4.1}--{S1.4.3}). Sensitivity maps of spatial checkpoints were employed as features for predicting the SA variables using cross-validated linear discriminant analysis (LDA). All units from a fixed spatially structured checkpoint were split into training and validation sets by applying the K-fold scheme ($k=5$) 100 times. Predictions were measured on the validation partitions, and the corresponding confusion matrices were aggregated by averaging. The resulting confusion matrices highlight linear separability patterns within SA variables. 

\subsubsection{Masked activation analysis}
\label{subsubsec:masked-actan}

Sensitivity analysis assigns a level of variability in activations in response to an applied augmentation. The resulting ``markup'' may be employed as a mask outlining more and less sensitive areas,
enabling guided editing of the activations.
Such perturbations break the communication pathways established among neurons from neighbouring layers.
It resembles the manual blocking (or activation) of sensitive receptors located at the synapse, changing the chemical context while leaving neuronal connections unchanged---i.e., a variant of reversible neurotransmission disruption, similar to the temporary effect caused by injecting a specific drug. The same idea stands behind the Dropout regularizing method, with the distinction that the latter completely zeroes out a subset of the output during the learning procedure.

Besides, there is a masking-based technique to estimate the sensitivity of occlusion attribution (e.g., \cite{petsiuk2018rise}), where masking is applied as a single transformation to identify specific areas of an input exerting the strongest influence on the prediction. Recall that the current work directly employs the intermediate changes of activations derived from different combinations of various input augmentations. Sensitivity values allow one to assess the contribution of every SA variable to the activations' variance. But how specific are the resulting sensitivity maps?

Figure~{\ref{fig:2-deep-sensitivity-examples}} demonstrates that sensitivity to several SA variables yields a compensatory effect for others. Moreover, there are intersections of augmentation actions, e.g., modifying the high-level feature like luminosity or sharpness. Therefore, the objective of the masked activation analysis is to assess the integral specificity of the sensitivity maps produced during simultaneous augmentation by matching them with single augmentations. Activation editing should not be confused with input augmentations: the latters are performed independently of any sensitivity maps, and activations within selected non-classification checkpoints are to be masked according to previously estimated sensitivity maps.

The current experimental series involves such editing to capture the changes in classification quality. The following masking variants were used: raw values, inverted values (1 - value), and thresholded versions of the raw values, $\text{thr}_{\alpha, q}(x) = \alpha x \cdot \mathbb{I}_{x \in q-\text{quant.}} + x \cdot \mathbb{I}_{x \notin q-\text{quant.}} $, where $\alpha \geq 0$ is a gain ($\alpha > 1$) or damping ($0 \leq \alpha < 1$) factor, $\mathbb{I}_{A}$ is an indicator of event $A$, $q-\text{quant.}$ is the $q$th quantile of all current sensitivity values at the corresponding checkpoint; in this work, $\alpha \in \{0, 0.5, 1.5\}$, $q \in \{0.5, 0.6, 0.7, 0.8, 0.9\}$.

The top-1 accuracy values measured for different masks form a new feature space for further analysis. These features were extracted for original inputs, as well as for those augmented by every single augmentation considered in the study. Then, the special correlation matrices were constructed as tables with rows corresponding to augmentations of input and columns representing SA variables used for masking (Figures~{\ref{fig:3-feature-corr-exp3_ilsvrc}}, \ref{fig:4-feature-corr-exp3_places365}). Each entry is a value of Spearman correlation corresponding to non-augmented input and input augmented by the same transformation used for masking. Further reasoning is based on the assumption that computed sensitivity values reflect the inherent sensitivity to the corresponding augmentation, and that matching the sources of distortion should lead to a greater deviation from the masked but non-augmented baseline, which may be reflected in a decrease in the corresponding correlation value. Since the last fully-connected layer decides the final prediction, it was excluded from the masking procedure (as well as other fully-connected checkpoints in the case of AlexNet and VGG11).

The sensitivity values were found to be skewed, exhibiting a non-uniform distribution, which prompted a further study to determine the single-class sensitivity patterns.
The first-order Sobol indices and Shapley values were adopted to address this problem. Total Sobol indices were excluded from the consideration due to their weaker specificity.
The candidate list of sensitive classes may be ranked by a simple sorting in descending order of the sensitivity values. However, the appropriate threshold for separating the non-sensitive classes is not clear. This work employed a top-5 approach as a heuristic guideline (Tables~{S1.7 (a)--(c)}). Using the same masking approach, a new pattern based on top-5 predictions was constructed to filter the results.
Predictions and sensitive classes were matched by computing the Jaccard index, which increases as predictions become more biased towards the sensitive classes (Figure~{\ref{fig:5-sobol-indices-jaccard}}, Tables~{S1.5.1}--{S1.5.2}).

\subsubsection{Cutting off the subnetworks and matching single-channel sensitivity on selected units}

In the last experimental series, standalone segments of neighbouring checkpoints were directly exposed to the input data.
As the number of input channels does not match those of the standard RGB,
original images were converted to HSV (see Figure~{\ref{fig:1-augmentation_examples}}), as the corresponding colour space splits higher-order features of the data into channels, and each new channel was separately repeated to fit the
corresponding segment (single-channel checkpoint). Selected subset of the most sensitive units (by means of at least one augmentation) was used to compute the corresponding sensitivity values. On top of them, the pixel-wise correlation patterns were extracted for HSV components, among the channels themselves and between each channel and the original values (Figures~{S1.8}). These experiments were conducted only on AlexNet and ResNet18. The higher-level tendencies were extracted by treating the correlations as new feature space and computing the higher-level correlations on their basis (Figures~{\ref{fig:6-mean-diff-corr-exp2}},~\ref{fig:7-mean-diff-corr-orig-exp2}).

\section{Results}
\label{sec:iden-sens}

Prediction accuracy for both augmentation sets, $A_1$ and $A_2$, constitute the baseline for comparison with accuracy on the non-augmented data, $A_0$ (Table~{\ref{table:0-accuracy_scores_on_augsets}}). These computations were performed on the validation partition; in the case of parameterized augmentations, a set of parameters was sampled three times for each data item. All the networks considered reveal that accuracy values change (decrease) unequally in response to a transformation. The most significant loss in accuracy is related to colour space alterations (hue, grayscale) and complete information removal (erasing). Augmentations that distort the input slightly have the lowest impact on the final score (horizontal flipping, local elliptic blurring, constant sharpness adjustment). These results were supplemented with an additional vision transformer, MaxViT-t~\citep{TuTZYMBL22}.

\begin{table}[!h]
\centering
\caption{Accuracy scores for selected networks measured on the validation partition of the corresponding datasets. All augmentations were applied independently as single transformations. For parameterized augmentations, each image was reused 3 times with different sets of sampled parameters. Underlined values are the scores most affected by the corresponding transforms.}
\label{table:0-accuracy_scores_on_augsets}
\caption*{Part A. Augmentation set 1.}
\vspace{-0.3cm}
\small
\input{./tables/0_accuracy_scores_part_A.tex}
\vspace{0.1cm}
\caption*{Part B. Augmentation set 2.}
\input{./tables/0_accuracy_scores_part_B.tex}
\end{table}

\subsection{Accessing neuronal activation sensitivity with Sobol indices and Shapley values}

The main results of this section are presented in Supplementary Figures~{S2}--{S3}.

\subsubsection{SA: Total variances}
\label{subsubsec:sa_totvar}

Variances and coefficients of variation reflect overall performance patterns measured at the checkpoints.
As clear from Figures~{S1.1.1}--{S1.1.5}, (a)--(b), distribution patterns for variances
are similar for both sampling schemes and vary depending on the augmentation set. Vision transformers demonstrate a greater similarity between variance maps on both augmentation sets (Figures~{S1.1.6}--{S1.1.7}, (a)--(b)).
Since a zero activation value corresponds to zero contribution during inference, the coefficient of variation (CoV), as the ratio of the standard deviation to the absolute value of the mean, was considered to compare units directly (Figures~{S1.1.1}--{S1.1.7}, (c)--(d)). They tend to be more uniform and further from zero, suggesting that most units are involved in inference. No connection can be drawn between the units most sensitive to a single augmentation and those with high variances
or CoVs (Figures~{S1.1}, {S2}--{S3}; Figure~{\ref{fig:2-deep-sensitivity-examples}}).

The exception is AlexNet trained on the Places365 dataset, where many neurons in the second, third and partially first checkpoints are inactive (Figures~{S1.1.4}). As the network was trained from scratch~\citep{zhou2017places} (see also the link in \hyperref[sec:data-avail]{Data availablility}), improper weight initialization could cause the outcome. This issue impacts all further steps of the analysis (hereafter, this remark is omitted), providing additional data on the framework's robustness. ResNet18 did not surface the problem, presumably due to its skip connection feature. Apart from that, similar tendencies as in ILSVRC may be observed.

\subsubsection{SA: Neuronal activation patterns}

\begin{figure}
    \hspace{-1 cm}
    \includegraphics[scale=1.2]{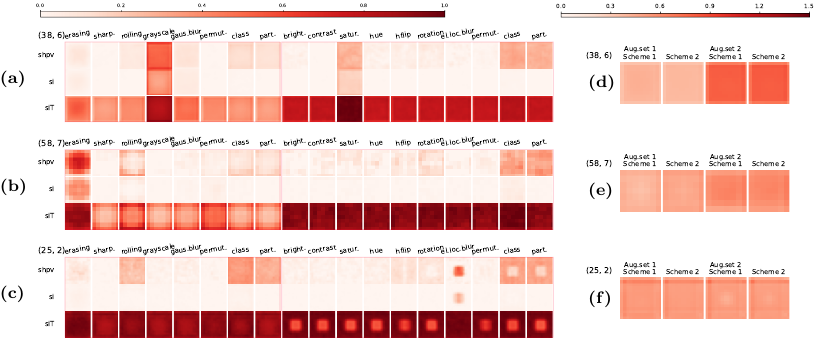}
    \caption{Examples of sensitivity to augmentations of the ILSVRC input manifesting in deep layers (a--c) and corresponding log-transformed coefficients of variation, CoV (d--f). (a), (d): AlexNet, checkpoint is the output from the features.7 module, $301$st output channel (neuron) sensitive to grayscale and saturation. Greater sensitivity to grayscaling is matched with a lower CoV value, and vice versa for saturation; (b), (e): VGG11, avgpool, $462$, erasing. The prominent sensitivity to erasing is not reflected in higher CoV values; (c), (f): ResNet18, layer3, $193$, rolling and elliptic local blur. For the latter transform, the sensitivity exhibits clear spatial structure, which is only slightly revealed in the CoV map.}
    \label{fig:2-deep-sensitivity-examples}
\end{figure}

In general, considered sensitivity values complement each other. For example, the first-order Sobol indices highlight sensitivity to a single variable,
and for some units it is hard to trace sensitivity to an elliptic local blur variable in this sense,
while total Sobol indices capture all its interactions and display the blob-affected patterns, for all other variables included in the augmentation set. Similarly, Shapley values capture the same interference resulting in elliptic blob patterns but are limited by a partition of unity constraint.
In addition, the elliptic local blur variable also confirms that sensitivity to input changes may last up to deeper layers. Such sensitivity
to transformations
may affect either all or certain neurons within a checkpoint, and the latter may support the existence of specialized structures. In Figure~{\ref{fig:2-deep-sensitivity-examples}}, (a)--(c), several examples of such deep neurons are provided. Surprisingly, sometimes sensitivity does not decrease (and even increases) with depth.

Similar trends hold for the vision transformers. At the same time, a number of their sensitivity maps additionally contain Greek cross and hash key patterns. As expected, separately plotted CLS checkpoints (ViT-b-16) show high sensitivity to the class variable, while the class-related sensitivity maps of patches (the spatial analogue of convolutional outputs) show a slight increase with depth, with a few patches exhibiting higher sensitivity values for the class variable at different checkpoints. The CLS parts share sensitivity to both the class and augmentation (colour adjustment, Gaussian blur) variables.

All networks show low sensitivity to permutation in terms of first-order Sobol indices and Shapley values, indicating that the constructed analysis mainly handles the augmentations themselves. As the corresponding total Sobol indices remain high, permutation appears to subtly affect the results through interactions with other variables (as indirectly confirmed by the second-order Sobol indices rendered in the source-code repository). The same interpretation may be applied to the partition variable, sensitivity to which is captured
by Shapley values (but neither first-order nor second-order Sobol indices): selection of a training-validation partition is essential to networks but in a higher-level sense.

All sensitivity values reflect sensitivity to the class variable. Its first-order Sobol indices are increasing while moving to deeper layers, reflecting that almost all network components are involved in classification and become increasingly specialized in target recognition with depth.

Standalone constant sharpness and horizontal flip augmentations weakly influence predictions (Table~{\ref{table:0-accuracy_scores_on_augsets}}) but demonstrate different sensitivity patterns during simultaneous augmentation: there is a low sensitivity for sharpness adjustment (Shapley values, first-order Sobol indices) at convolutional checkpoints. In contrast, the horizontal flip has structured sensitivity patterns (Shapley values).

Rolling generates either border bar patterns (first-order Sobol) or more uniform full-cover sensitivity involving interactions (Shapley values). Local patterns appear for erasing and elliptic local blur augmentations (first-order Sobol, Shapley). Erasing affects both predictions and activations severely, i.e., networks more effectively compensate for elliptic blurring as a local transform preserving more information.

All networks demonstrated strong sensitivity to grayscaling. As might be expected, this sensitivity intersects with saturation sensitivity for specific units since grayscale is a limiting case of this transform. A hue transform is another colour adjustment that strongly impacts both sensitivity and accuracy, indicating the importance of a natural colour palette.

\subsubsection{Correlation patterns}
\label{sssec:cor-patterns}

Correlation matrices between SA variables are given in Figures~{S1.2.1 (a)--(e), S1.2.2 (a)--(b), S1.2.3 (a)--(c)}.

Correlations within the augmentation set tend towards a common level as checkpoints progress deeper, clustering variables together and smoothing their impact. In the case of ViT-b-16, however, the results differ for patch-based and CLS-based parts: while the CLS part follows the mentioned trend, the patch-based correlations preserve their values regardless of the depth.

Grayscale has a strong correlation with saturation and hue by means of the first-order Sobol indices and Shapley values of CNNs, Swin-t and ViT-b-16 (patches). For the beginning checkpoints, it tends to correlate more with saturation, giving way to a correlation with hue in the deeper layers.
In both cases, these correlations decay to zero as checkpoints move closer to the avgpool (however, AlexNet breaks the pattern exactly at this checkpoint). Hue and saturation variables are also correlated, and their correlation decays to zero or their cluster's common level (except ViT-b-16, CLS). Other notable positive correlations are observed in pairs from $\{ \text{hflip}, \text{rolling}, \text{rotation} \}$, in terms of Shapley values.

Total Sobol indices distinguish the erasing transform as significantly separated from other augmentations (except ViT-b-16, CLS), forming a distinct cluster at several checkpoints closer to the beginning of a network. A similar trend to cluster separately is slightly less pronounced for the hue variable, and even less prominent for grayscale and saturation.

\subsubsection{Spatial correlation maps}

Spatial correlation maps relating the SA variables and CoVs can be found in Supplementary materials as Figures~{S1.3.1 (a)--(c), S1.3.2 (a)--(b), S1.3.3 (a)--(c)}.

The class variable has near-zero correlations for total Sobol indices
measured at the final (classification) checkpoint. In addition, for single augmentations
the final-layer sensitivity
shares either the inhibition pattern or near-zero correlations, suggesting that the activations lose augmentation-explained variability there.
Correlation patterns based on Shapley values for the last layer are more diverse than ones for Sobol indices, and the common trend for all considered networks is the same near-zero or weak positive correlation for the class variable and a weak negative correlation for the partition variable.

For convolutional and patch-based transformer checkpoints, these correlation maps are non-scalar and may be viewed as spatial on-off patterns similar to ones in biological networks: positive correlation areas reflect the level of alignment between sensitivity and activation, corresponding to excitation patterns, while negative correlations can be related to inhibition patterns. And similarly to biological networks~\citep{rossi2020spatial,tring2022off}, the corresponding receptive fields demonstrate both center-surround and distributed structures. These spatial patterns depend on network, dataset, checkpoint, and SA variable, as well as on sensitivity values. Distributed patterns are minor for total Sobol indices, while the first-order Sobol indices and Shapley values exhibit both. As in biological neurons, the on-off fluctuations exist~\citep{liang2018distinct}: for a fixed network, the class or partition spatial correlation maps are similar in general but locally different depending on the augmentation set.

Certain patterns that emerged in vision transformer-related maps resemble Gabor wavelets.

\begin{figure}[ht!]
    \vspace{-1.5 cm}
    \centering
    \includegraphics[scale=0.9]{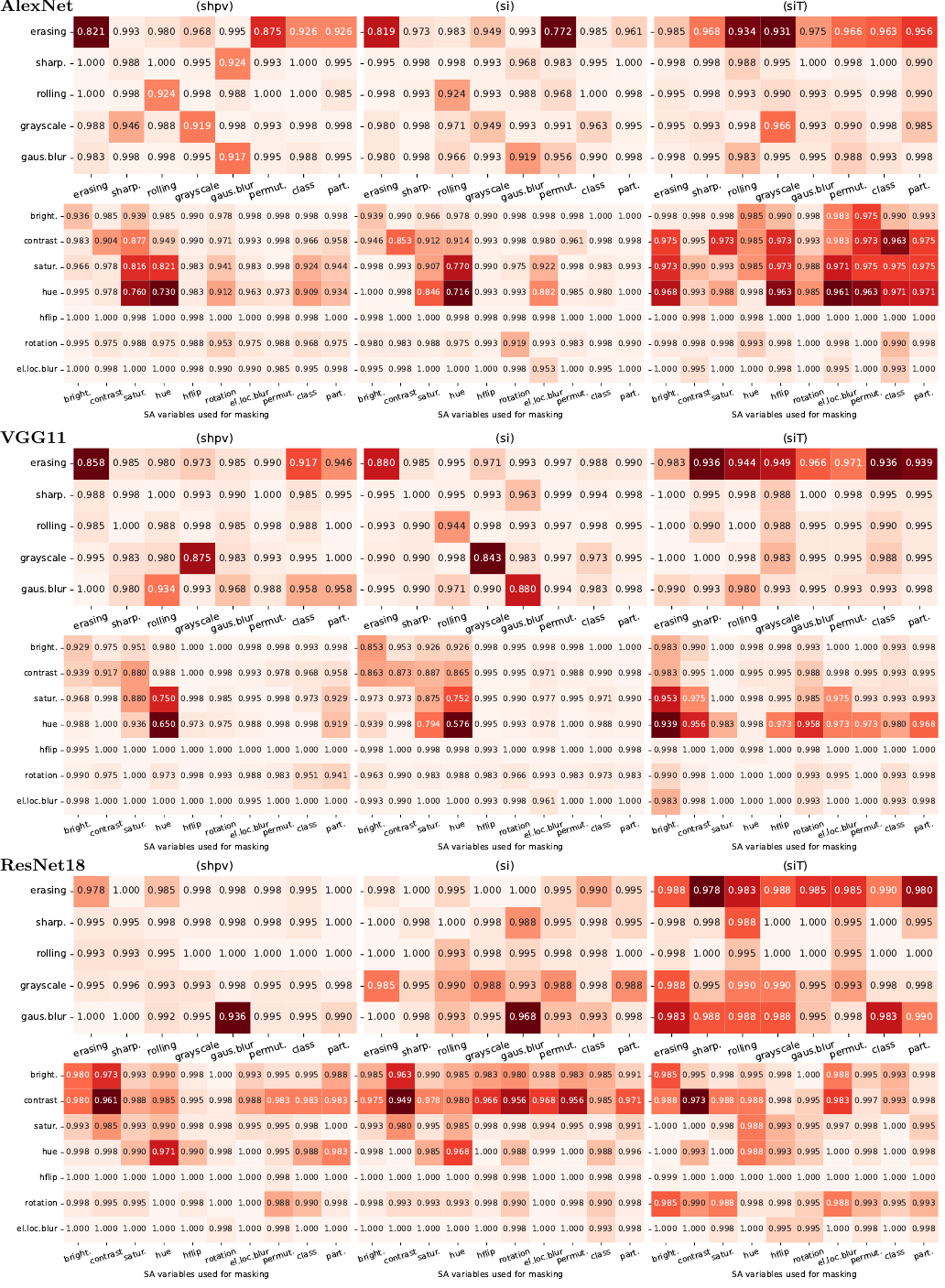} 
    \caption{(ILSVRC) Correlation patterns extracted by guided damaging of activations during inference. Darker colours correspond to lower correlation values. Each item within a heatmap is the Spearman correlation that relates augmented and non-augmented masked inputs. Rows correspond to augmentations used for transforming the input, columns indicate the SA variables used as sources for building masks. Sensitivity values are designated as follows: \textit{shpv} for Shapley values, and \textit{si} and \textit{siT} for the first-order and total Sobol indices, respectively.}
    \label{fig:3-feature-corr-exp3_ilsvrc}
\end{figure}
\FloatBarrier

\begin{figure}[ht!]
    \centering
    \includegraphics[scale=0.9]{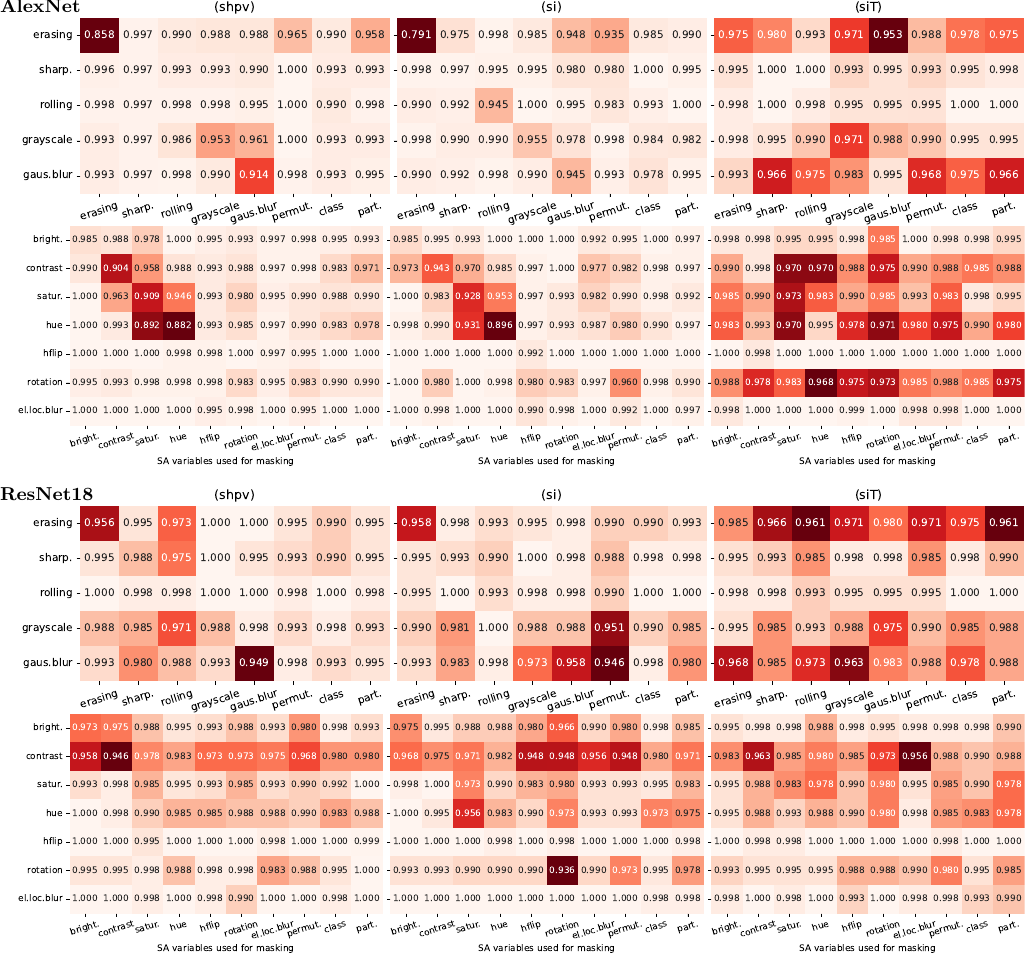} 
    \caption{(Places365) Correlation patterns extracted by guided damaging of activations during inference. Darker colours correspond to lower correlation values. Each item within a heatmap is the Spearman correlation that relates augmented and non-augmented masked inputs. Rows correspond to augmentations used for transforming the input, columns indicate the SA variables used as sources for building masks. Sensitivity values are designated as follows: \textit{shpv} for Shapley values, and \textit{si} and \textit{siT} for the first-order and total Sobol indices, respectively.}
    \label{fig:4-feature-corr-exp3_places365}
\end{figure}

\subsubsection{Linear discriminant analysis}
\label{sec:lda-main}

Confusion matrices derived from cross-validated LDA are presented in Figures~{S1.4.1 (a)--(c), S1.4.2 (a)--(b), S1.4.3 (a)--(b)}. These results are detailed in~{\ref{appendix:lda}}; the following paragraphs summarize the common tendencies.

SA variables are well separated at the first checkpoint for all sensitivity values and convolutional networks, except for AlexNet trained on the Places365 dataset (see remark in Section~\ref{subsubsec:sa_totvar}). This separability decays while moving deeper, which is consistent with previously identified correlational patterns. Strong separation of the erasing transformation persists even at the deepest checkpoints. A possible explanation for this pattern is that it causes severe distortions, which are difficult to compensate for and can be easily detected even by a linear classifier. Additionally, the closer a checkpoint is to the classification layer, the more separable the class variable becomes---exhibiting either U-shaped or nearly parallel trends, along with poorer separability for certain augmentation variables. This may stem from the cancellation of augmentation impact and the retargeting of neurons to the class variable, as the network complies with the inference goal and learns higher-level abstractions.

A distinctive feature of vision transformers is the higher separability of SA variables with a lower degree of cross-confounding. Swin-t, however, exhibits the same pattern of decaying separability as CNNs, which can be attributed to its architecture. Swin-t produces a multi-scale feature hierarchy that is similar to the principle of CNN architecture design, while ViT-b-16 preserves spatial dimensionality in its dataflow without downsampling and thus does not share the trend.

\subsection{Pattern analysis for damaged activations}

The following results were obtained by applying guided masking to activations from convolutional checkpoints during inference.

\subsubsection{Correlation patterns}

Figures~{\ref{fig:3-feature-corr-exp3_ilsvrc}},~\ref{fig:4-feature-corr-exp3_places365} present the matching results for input augmentations and guided activation masking. Certain pairs for Shapley values and first-order Sobol indices are the most selective: Gaussian blur, grayscale, erasing, and hue transforms are the most diverging when masking and augmentation variables coincide,
resulting in
lower diagonal values. It supports the matching between computed sensitivity values and inherent network sensitivity. Total Sobol indices break this pattern in most cases (except grayscale and a few other partial entries), presumably due to their structure incorporating higher-level interactions among SA variables.

On the first augmentation set, the erasing transform is well-separated in terms of Shapley values and Sobol indices (exceptions: AlexNet, the erasing-permutation pair dominates erasing-erasing on the ImageNet dataset; ResNet18 - all values are high but still separable on Places365); for total Sobol indices, entries deviate significantly across all pairs. Grayscaling is separable in the case of AlexNet and VGG11 for all sensitivity values, whereas for ResNet18, the entries are either close to one (ImageNet, Shapley) or diverging. The Gaussian blur's matching is rarely impaired (VGG11, Shapley values; all networks, total Sobol indices). The rolling is well-discriminated for Sobol indices (except for ResNet18, Places365) and for AlexNet's Shapley values (ImageNet).

On the second augmentation set, AlexNet's and VGG11's results are aligned for the ImageNet dataset: both networks
showed
selective sensitivity to
elliptical local blurring and rotation according to Sobol indices. The horizontal flipping resulted in a correlation coefficient of nearly one for all sensitivity values, networks, and datasets. The remaining colour adjustments are highly cohesive.
The brightness is separable in many cases (even for total Sobol indices, VGG11 and ResNet18, ImageNet), but is susceptible to confounding with other SA variables in the case of ResNet18 (Shapley values, the first-order Sobol indices). The hue adjustment pairs well with the same masking SA variable, although partially overlapped with the saturation variable.
The saturation itself is mostly entangled with hue and slightly with contrast (ResNet18, ImageNet; AlexNet, Places365). The contrast transformation matches well with its masking for all ResNet18's sensitivity values (with the only exception, Sobol indices on Places365), though it is highly affected by other SA variables (especially other colour adjustments) and is entangled with saturation for AlexNet and VGG11 (Shapley values).

In addition, for erasing, hue, contrast, Gaussian blur, and grayscale augmentations, masking with other variables significantly affects predictions, indicating possible instabilities related to these augmentations. This view is supported by the baseline measurements given in Table~{\ref{table:0-accuracy_scores_on_augsets}}, as well as by the patterns described in the previous subsection. It may also indicate intersections among sensitivity maps, and the more localized it is, the weaker the effect on prediction patterns (for instance, horizontal flipping, local elliptic blur; erasing is less localized, especially in deeper checkpoints). Although sharpness adjustment by a constant factor has little effect on regular predictions, it is a global augmentation; hence, other variables influence its prediction pattern. In fact, on ImageNet, it is affected mainly by the Gaussian blur variable, which operates on the same abstract property (sharpness). As for Places365 (ResNet18), this pattern is reversed: the blurred input tends to be affected by the sharpness variable.

The established findings were also redisplayed in the hierarchical clustering analysis (HCA) ({\ref{appendix:hca}}).

\begin{figure}[t]
\small
\begin{subfigure}[h]{\linewidth}
\centering
\caption{Imagenet (ILSVRC)}
\includegraphics[scale=0.45]{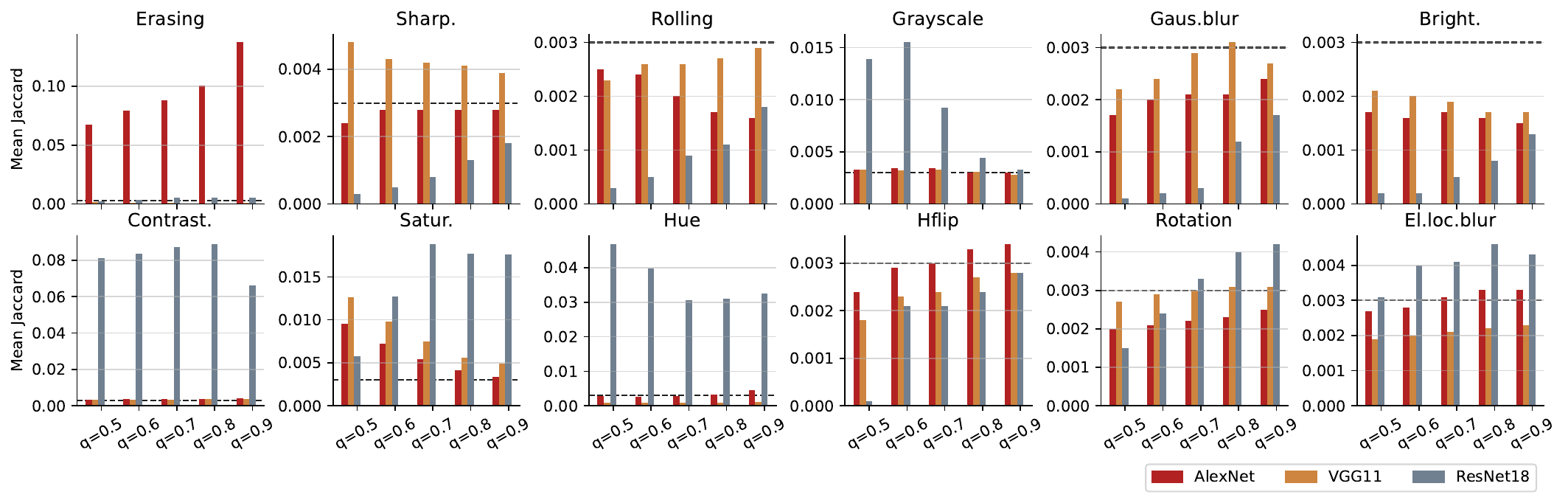}
\end{subfigure}
\begin{subfigure}[h]{\linewidth}
\centering
\caption{Places 365}
\includegraphics[scale=0.45]{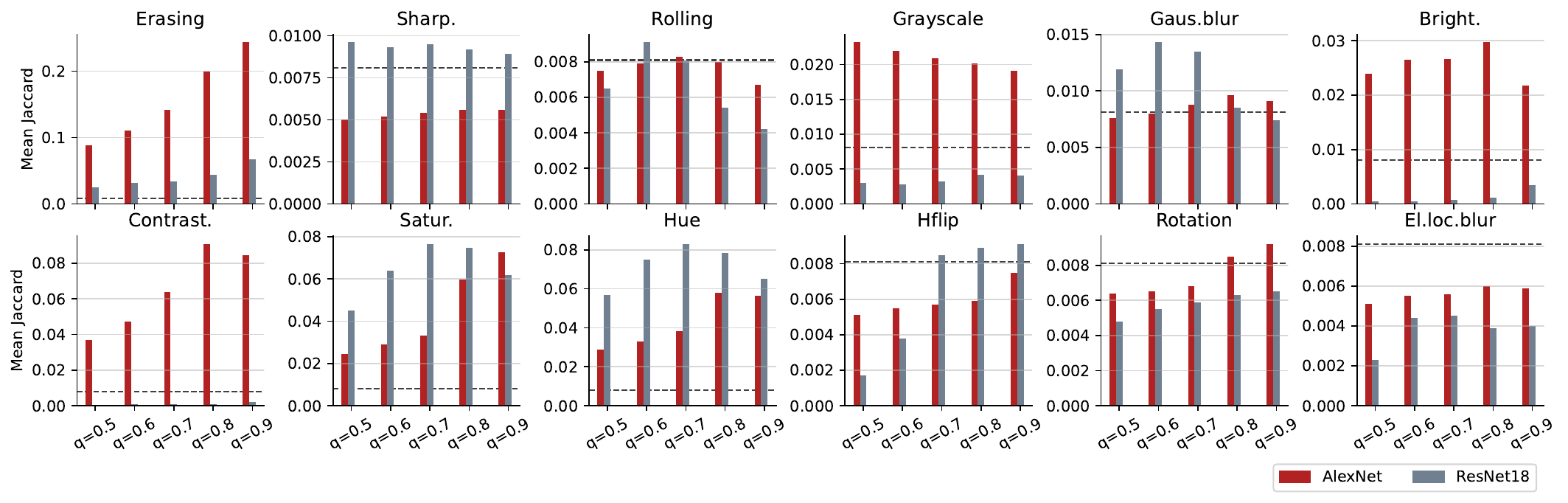}
\end{subfigure}
\caption{Mean Jaccard indices between top-5 masked predictions and top-5 sensitive classes extracted using Sobol indices computed for the final (classification) layer. The sensitivity analysis variable used for masking (convolutional checkpoints only) matches the input augmentation. Activations corresponding to the $(1-q) \cdot 100\%$ top sensitivity values per checkpoint were multiplied by $\alpha=1.5$ during inference. Dashed lines indicate the interquantile-based thresholds $\tau$.}
\label{fig:5-sobol-indices-jaccard}
\end{figure}

\subsubsection{Single-class sensitivity to augmentations}

Tables~{S1.7 (a)--(c)} list the top-5 classes most sensitive to considered augmentations according to conducted sensitivity analysis. Corresponding selectivity patterns cannot be reliably deduced from divergent values only. Thus, these sensitive classes were matched with the top-5 masked predictions using the Jaccard index (Tables~{S1.5.1 (a)--(g), S1.5.2 (a)--(g)}) to highlight outliers according to the precomputed threshold values (see~{\ref{appendix:thresh_scsa}}). Figure~{\ref{fig:5-sobol-indices-jaccard}}, based on Tables~{S1.5.1 (c), S1.5.2 (c)}, is provided separately since it displays the most prominent evidence of proper sensitive class selection manifested as a substantial increase in Jaccard indices for the erasing augmentation applied to AlexNet's input. The objects of the corresponding classes (\textit{screen}; \textit{monitor}; \textit{television}; see Table~{S1.7 (b)}) contain a filled rectangular pattern, similar to the results of applying the erasing transform. Other notable examples include classes that contain
light-shadow or edge contrast and are presumed to be sensitive to contrast adjustment (\textit{cliff dwelling}; \textit{jigsaw puzzle}).
A detailed analysis of other classes and their augmentation sensitivity is provided in~{\ref{appendix:scsa}}.

\nopagebreak[0]
\subsection{Cutting off submodules of a network}

\begin{figure}[t]
    \centering
    \includegraphics[scale=0.9]{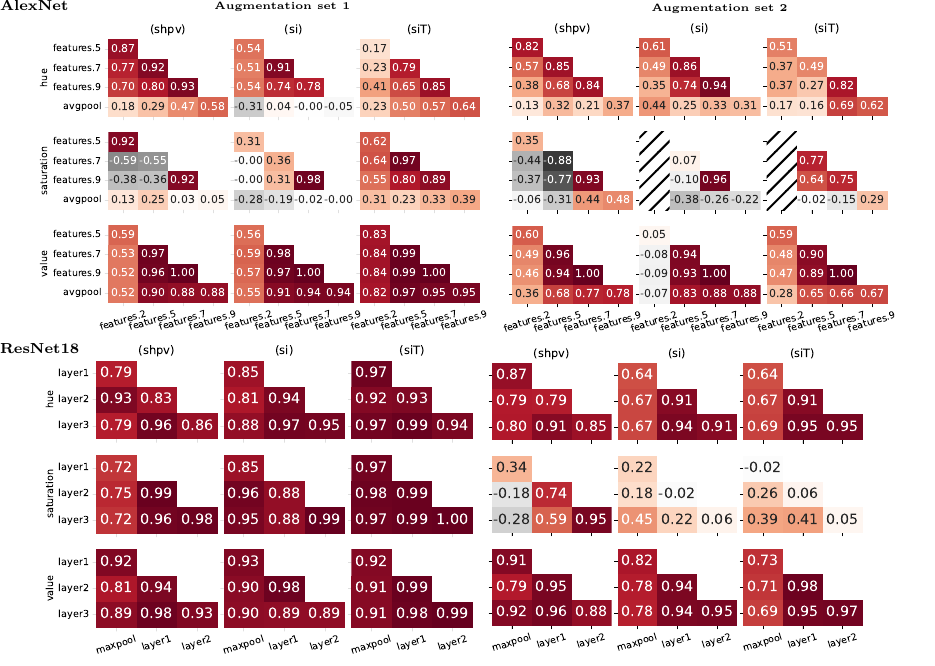}
    \caption{Higher-order correlation patterns extracted for segments separated from their networks: paired checkpoints within a fixed input channel (isolated segments only).  Single heatmap legend: rows correspond to single-channel checkpoints, columns to those of the original. Lighter colours encode correlations closer to zero (in their absolute value). Hatched cells indicate constant input that could not be correlated.}
    \label{fig:6-mean-diff-corr-exp2}
    \vspace{-0.15 cm}
\end{figure}

\begin{figure}[t]
    \centering
    \includegraphics[scale=1.1]{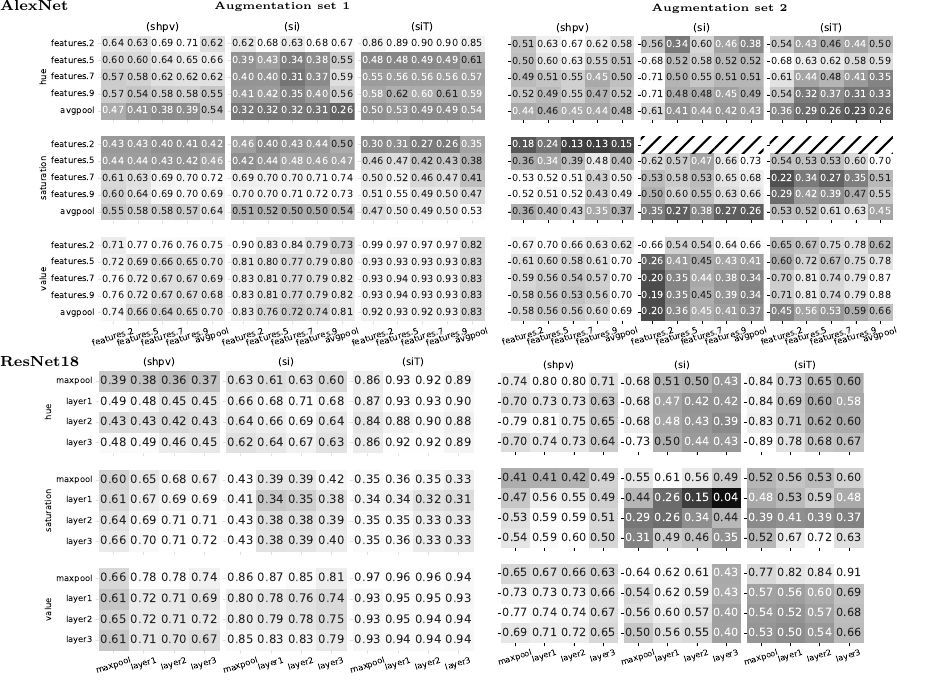}
    \caption{Higher-order correlation patterns extracted for segments separated from their networks: matching the isolated and original segments. Single heatmap legend: rows correspond to single-channel checkpoints, columns to those of the original. Darker colours represent lower values. Hatched cells indicate constant input that could not be correlated.}
    \label{fig:7-mean-diff-corr-orig-exp2}
    \vspace{-0.15 cm}
\end{figure}

First, Spearman correlations between paired SA variables were computed at the checkpoint level for both isolated fixed-channel segments (hue, saturation, value channels) and the corresponding original parts. Corresponding unitwise-averaged correlation matrices
are presented in Figures~{S1.8 (a)--(d)} (AlexNet and ResNet18 on ImageNet only). They were further processed to extract higher-level correlation patterns between single-channel segments: cross-correlations between the triangular parts of these matrices were computed, matching different checkpoints within a channel (Figure~{\ref{fig:6-mean-diff-corr-exp2}}) and relating standalone outputs to the original flow (Figure~{\ref{fig:7-mean-diff-corr-orig-exp2}}).

As for AlexNet, the value and saturation components are the most and least correlated with the original, respectively. Its
successive
checkpoints also tend to be more correlated, which could be
due to
their coherence
emerged through training. ResNet18 shares the same channel ordering but exhibits a higher correlation level
irrespective of checkpoint distances, which can be explained by skip connections that lead to longer retention of distortions.
The most diverging correlations are observed at checkpoints close to the input (both networks) and to the fully connected layers (AlexNet). The first pattern may reflect misalignment between standalone segments, tuned to specific channels and value ranges, and the input data directly passed to these segments.
The second case might have arisen due to the suppression of spatial details by adaptive average pooling: this layer reduces spatial dimensionality and minimizes overlapping regions, resulting in data smoothing that merges local patterns.

For channel-original matching, there is no diagonal dominance and, consequently, no exclusive similarity between isolated and non-isolated segments of the same checkpoints. They match in different positions despite input data differences, which could mean that sensitivity patterns are translated throughout the entire convolutional part of a network to a reasonable extent. There is a decrease in some correlations for distant original checkpoints, which can be explained by the gradual accumulation of discrepancies.

\section{Error analysis}
\label{sec:error-analysis}

\subsection{Sobol indices and Shapley values assessment}
\label{sec:si-shpv-assessment}

The Sobol-Saltelli method for computing Sobol indices and the Monte Carlo scheme for Shapley values output approximate results, which may violate the theoretical properties of these quantities. Among the computed Sobol indices, no violations of the inequality $S_i \leq S_i^T$ were observed. However, there are first-order Sobol indices and Shapley values that have dropped below zero, and several total Sobol indices that have exceeded the upper limit, motivating the necessity of conducting error analysis. Since each checkpoint produces a bundle of interconnected activations affected by similar external impacts, deviations from the theoretical bounds contain information regarding numerical noise and computational convergence. In addition to the sensitivity values, estimations of means and variances are available in this study. Even though these quantities are independently computed approximations, they reduce the degree of uncertainty and contribute to the error analysis.

To construct the appropriate function for error analysis, the variances corresponding to values that drop outside the theoretical bounds are to be selected and normalized by the variance range, that is, the dynamic range of variance across all outputs. Since there are multiple activations, they are to be aggregated by their maximum value 
in order to represent the worst-case scenario as a scalar, dimensionless value. An additional

\begin{figure}[h!]
    \hspace{-0.75 cm}
    \includegraphics[scale=0.35]{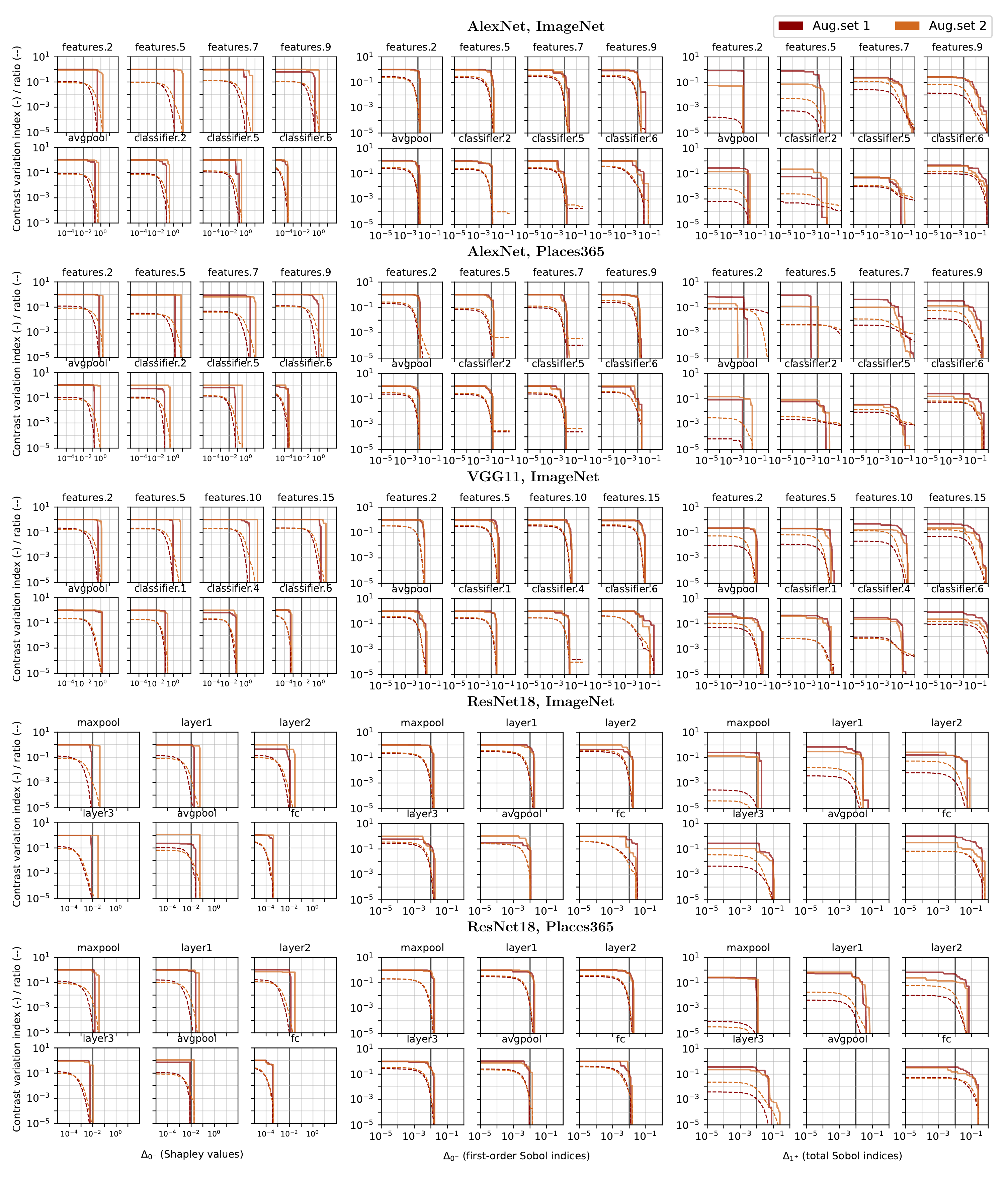}
    \caption{(CNNs) Assessment of the sensitivity values from the first experimental series.}
    \label{fig:8-errplots-cnns}
    \vspace{-0.15 cm}
\end{figure}
\FloatBarrier

\noindent dependency on the discrepancy value completes the construction of the contrast variation index:
\begin{equation}
\label{eq:contrast-variation-index}
    \text{CVI}(\Delta; F, A, \rho_{\pm}) = \frac{\max\limits_{x \in D(F), k=\overline{1,K}} \Big\{ V[F(x)] \, \Big| \, \rho_{\pm}(V_k[F(x)], A \Big. ) > \Delta \Big\}}{\max_{x \in D(F)} V[F(x)] - \min_{x \in D(F)} V[F(x)]},
\end{equation}
where $F(x)$ is a mapping to be investigated, $D(F)$ is its domain, $V[\cdot]$ is a variance, $V_k[\cdot]$, $k=\overline{1, K}$ are sensitivity values corresponding to $K$ SA variables, $A$ is a value that constitutes the theoretical bounds, $\Delta$ is a discrepancy value, and $\rho_{\pm}$ is either $\rho_{+}(x, y) = \max(x - y, 0)$ or $\rho_{-}(x, y) = |\min(x - y, 0)|$.
The index is not applicable to the isotropic case with equal or minimally differing activation variances, as its denominator becomes equal to or approaches zero. It may occur when a checkpoint ends with a trivial layer with equal weights, an untrained layer without symmetry breaking applied to its weights, or a layer trained with activation covariance-based regularization. Another possible example is a BatchNorm layer with activation-wise statistics computed over the batch dimension, when the effect of augmentations applied to the input is not sufficiently distinct. To increase the robustness of this quantity, it can also be modified by replacing minimums and maximums with quantiles. The current study adheres to a conservative approach without such modifications.

\begin{figure}[t]
    \hspace{-0.75 cm}
    \includegraphics[scale=0.35]{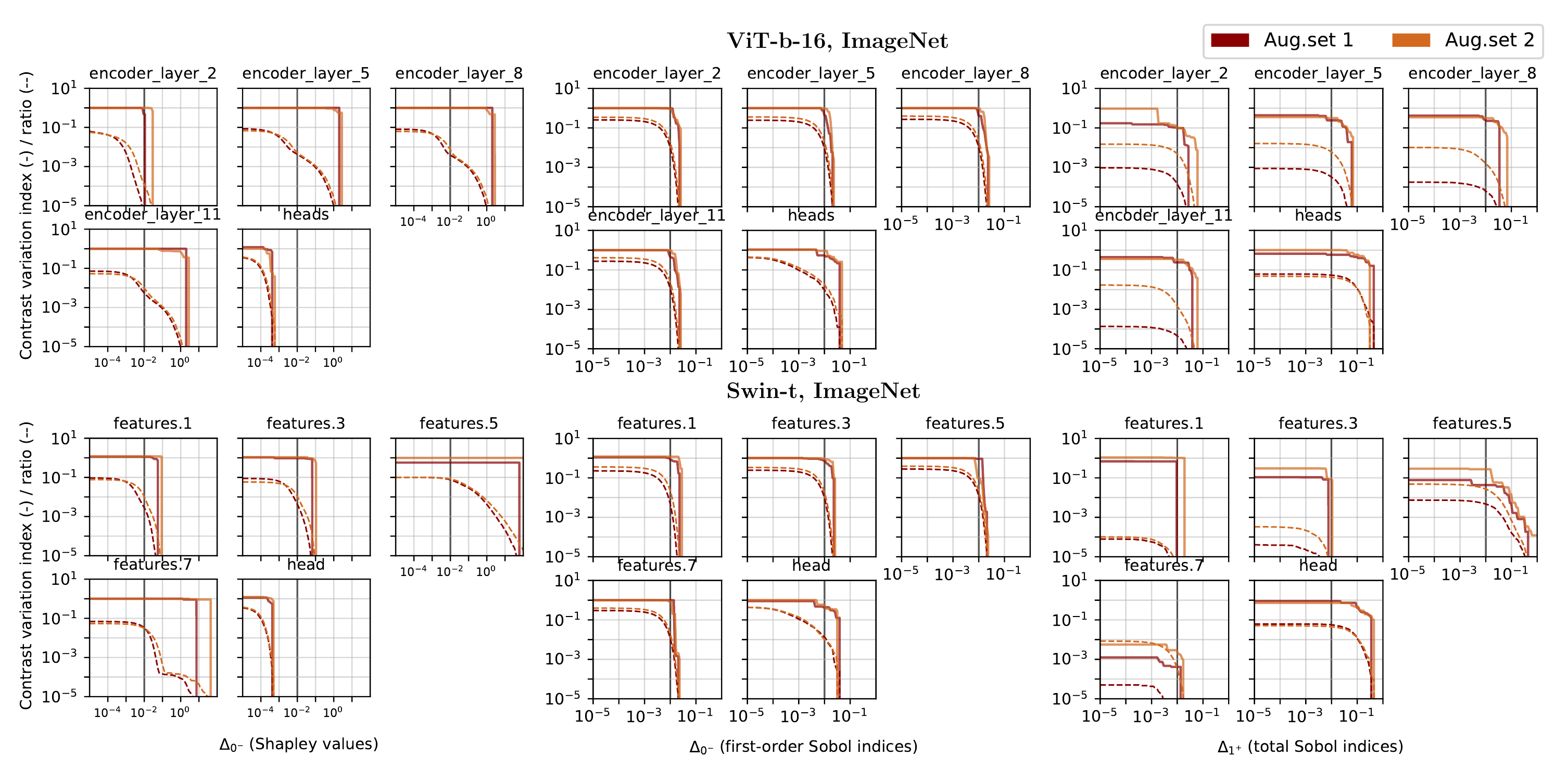}
    \caption{(Vision transformers) Assessment of the sensitivity values from the first experimental series.}
    \label{fig:9-errplots-vt}
    \vspace{-0.15 cm}
\end{figure}

The constructed index is to be used for error analysis as follows: with fixed $F$, $A$, and $\rho_{\pm}$, the contrast variation index is computed for different values of the threshold $\Delta \equiv \Delta_{A^{\pm}}$. The lower the index, the less influence the errors exert on the sensitivity analysis, since in practice, computed values are clipped to conform to their theoretical bounds. Another important characteristic is a fraction of erroneous values among all outputs, and the dependence of this fraction on the same threshold enables one to assess the size of the effect. The summary results of the error analysis for the sensitivity values obtained in the first experimental series are presented in Figure~{\ref{fig:8-errplots-cnns}} (CNNs) and Figure~{\ref{fig:9-errplots-vt}} (vision transformers).

As the widely accepted threshold for distinguishing sensitivity and insensitivity of the SA variable is $10^{-2}$~\citep{SARRAZIN2016135}, it therefore appears reasonable to use the same value as a reference within the scope of the analysis.
The first-order Sobol indices require the use of the following parameters: $\rho_{-}$, $A = 0$, and hence $\Delta \equiv \Delta_{0^-}$. For most checkpoints of AlexNet, both the contrast index and the miscomputed rate drop significantly as they approach the threshold. The exceptions are the final classification checkpoints, which preserve the higher index values for larger discrepancies, though the rate still drops considerably. Similar tendencies are observed at features.7 (both datasets) and feature.9 (ImageNet) checkpoints. ResNet18 shows stabilizing patterns similar to AlexNet. As for VGG11, the drop region is notably shifted towards $\Delta_{0^-} = 10^{-1}$ at all checkpoints. In theory, the sample size could be increased in an attempt to stabilize the results. However, due to its architectural specifics, this network is expensive in terms of computations, inference, and storage of the results.

Total Sobol indices employ $\rho_{+}$, $A = 1$, and $\Delta \equiv \Delta_{1^+}$ for error analysis. Despite exhibiting similar trends in terms of discrepancy size rate, the total Sobol indices stabilize more slowly and deviate from the theoretical bounds to a greater extent, with distances approaching  $10^{-1}$ or even larger. The size rate for the latter drops significantly below $0.1$, suggesting less impact of the extreme discrepancies on the results. Nevertheless, these values are of limited use within the framework.

Computed Shapley values are unnormalized, non-negative quantities, thus for error analysis, one may use parameters similar to those for first-order Sobol indices. However, the threshold $\Delta = 10^{-2}$ becomes too restrictive for variance values greater than one. The corresponding discrepancies exert less influence on the results. The ResNet18 errors are stabilizing within the $[10^{-2}, 10^{-1}]$ range, tending towards the left boundary of the interval. For AlexNet and VGG11, the dropping region of the contrast variance index shifts to $[10^{-1}, 10^{0}]$, while the rate of the errors starts to decrease around $\Delta_{0^{-}} = 10^{-2}$. An important observation is that all the classification checkpoints are affected by the errors to a lesser extent, which supports the use of Shapley values for single-class sensitivity analysis.

Total Sobol indices display the most unstable results at the classification layers. Around $10\%$ of them preserve the high values of contrast variance index at $\Delta_{1^+} \sim 10^{-2}$. The influence of such instability is minimal, since the single-class sensitivity analysis is built on top of first-order Sobol indices and Shapley values only.

In this regard, it is worth discussing the classification layers themselves. 
In this study, the softmax outputs were used, motivated by the objectives of the single-class sensitivity analysis of decision-making systems that may rely on probability estimates. Despite additional dependencies introduced by the softmax, it provides insights into prediction confidence. 
As its variance is bounded, these insights could be utilized for distinguishing the low-variance classes: confident predictions (mean value significantly distant from zero) and confident rejections (near-zero mean value). On the basis of changes in confidence rankings caused by guided masking, one could construct additional criteria for single-class sensitivity analysis, in particular to address possible instability issues regarding first-order Sobol indices.

The discovered dependencies are also applicable to vision transformer networks.
\FloatBarrier
\subsection{Average correlations}
\label{sec:avg-corr}

The conducted correlation analysis (Section~\ref{sssec:cor-patterns}) partially involves averaging of the correlation matrices. The variance-stabilizing Fisher transform $\text{arctanh}(x)$ was applied before averaging, followed by its inverse after computations. 
To assess the robustness of the results, two-sided 95\% conﬁdence intervals ($N = 1000$) were computed using bootstrapping, by sampling with replacement from the set of whole correlation matrices precomputed for each spatial point. 
Similarly to the assessment of sensitivity values, the maximum statistic was employed.

In CNNs, the widths tend to increase with depth, except for the (AlexNet, Places365) pair, reaching their maximums at the last spatial checkpoint (avgpool or layer3). The maximum widths of their confidence intervals for Shapley values, first-order Sobol indices, and total Sobol indices, respectively, are: for (AlexNet, ILSVRC), $0.084$, $0.096$, and $0.113$; for (AlexNet, Places365), $0.145$, $0.086$, and $0.119$; for (VGG11, ILSVRC), $0.044$, $0.048$, and $0.069$; for (ResNet18, ILSVRC), $0.041$, $0.038$, and $0.066$; and for (ResNet18, Places365), $0.038$, $0.037$, and $0.075$.

As for vision transformers, no clear trend is observed. Several worst cases reach moderate values, remaining suitable for the qualitative analysis performed. The corresponding maximum widths of the confidence intervals (same order as above) are: for (ViT-b-16, ILSVRC), $0.092$ (encoder layer 5), $0.042$ (encoder layer 5), and $0.105$ (encoder layer 8); and for (Swin-t, ILSVRC), $0.181$ (features.5), $0.06$ (features.7), and $0.143$ (features.5).

\subsection{Linear discriminant analysis}
\label{sec:lda-err}

The LDA experiments (Section~\ref{sec:lda-main}) include 100 repetitions of a stratified 5-fold cross-validation scheme (with balanced classes, as each SA variable has an equal number of units). The reported results are averaged confusion matrices measured on the validation folds. 
Their standard deviations can be used to assess the stability of the results, showing how well the averages are aligned. Their maximum values within the confusion matrices are reported below.

In the (AlexNet, ILSVRC) pair, the maximum standard deviations are observed at the first checkpoint, $0.186$, $0.142$, and $0.145$ (Shapley values, first-order Sobol indices, total Sobol indices), presumably indicating the influence of input data variability, higher spatial dimensionality, and lower number of samples 
(spatial maps). 
At deeper checkpoints, the maximum standard deviation tends to be reduced and is bounded by $0.085$, $0.099$, and $0.082$ (same order as above), which points to the stability of the results. For (Alexnet, Places365), high maximum values are observed up to the third checkpoint: $0.353$, $0.321$, and $0.334$ (same order as above), due to the large number of non-performing units. The bounds for the remaining two checkpoints drop significantly: $0.088$, $0.095$, and $0.069$ (same order as above), presumably reflecting a compensatory effect during inference.

ResNet18 displays a similar decreasing trend, but starting from the layer.1, which can be explained by the specifics of its first checkpoint (maxpool). The maximum standard deviation values (same order as above) are: for (ILSVRC, the first three checkpoints), $0.271$, $0.155$, and $0.151$; for (ILSVRC, the last spatial checkpoint), $0.071$, $0.073$, and $0.067$; for (Places365, the first three checkpoints), $0.275$, $0.153$, and $0.149$; and for (Places365, the last spatial checkpoint), $0.074$, $0.071$, and $0.069$.

VGG11 on ILSVRC also follows the same decreasing trend: at features.2, $0.147$, $0.126$, and $0.093$; at features.5, $0.16$, $0.101$, and $0.075$; at features.10  and features.15, $0.064$, $0.077$, and $0.071$; and at avgpool, $0.066$, $0.057$, and $0.05$ (same order as above).

Vision transformers exhibit stable levels of maximum standard deviations across all considered checkpoints, with the only exception being Swin-t, which shows relatively higher bounds for Shapley value-based results at the first checkpoint ($0.155$;  $0.038$ - at the others). The corresponding maximum standard deviation values are: for (ViT-b-16, ILSVRC), $0.031$, $0.043$, and $0.041$; and for (Swin-t, ILSVRC), $0.155$, $0.08$, and $0.077$ (same order as above).

\section{Discussion}
\label{sec:disc}

This work aimed
to design a framework for studying neural networks by exploring the sensitivity of their internal activations to simultaneously applied augmentations, enriched with auxiliary variables. Sobol indices and Shapley values were used as the primary computational tools, both of which require appropriate sampling schemes to apply augmentations simultaneously. Resulting sensitivity maps provide a visualization which captures the individual sensitivity of single units within the checkpoint. 

\textbf{Models and datasets assessment}. For a long time, ImageNet has been the leading dataset in deep learning, and it remains a pretraining basis for various computer vision models. In this research, leveraging ILSVRC (as a subset of ImageNet for training partition and independently sampled for validation set), the developed framework has been thoroughly evaluated on three classic CNN architectures, AlexNet, VGG11, and ResNet18, and supplemented with additional experiments on two vision transformer networks, ViT-b-16 and Swin-t. Table~{\ref{table:0-accuracy_scores_on_augsets}} indicates that all of them still exhibit a non-negligible validation error, suggesting that they are unlikely to have surpassed the dataset limitations. Additionally, sensitivity patterns of AlexNet and ResNet18 were investigated on the Places365 dataset, similar to ILSVRC in encompassing the class and imaging conditions diversity, offering a sufficiently high resolution, and featuring accessible pretrained models.

\textbf{Image classifiers extensively rely on colour features}. All networks experience the largest drop in performance when the colour space modifications are applied to the input. In the first place, it is a hue adjustment, followed by saturation, and grayscaling as its limiting case. In many evaluations, they are difficult to distinguish, as are adjustments to brightness and contrast.

\textbf{Sensitivity to input augmentations propagates through the entire network}. Notably, the sensitivity to quite simple augmentations is observed up to the deepest layers, which is consistent with the findings evaluated in the work~\citep{qin2018convolutional}, demonstrating low-level concept identification
by deep neurons. Even the deeper layers of high-performing networks may respond to the input damage. Presumably, the networks learned to compensate for these inclusions, leaving a potential direction for adversarial attacks, if the data flow can be appropriately modified. In the simplest case, one can inhibit or highlight the sensitive activations using masks. Another important point to consider is the analysis of total variances and coefficients of variation, as their high values may also indicate instability. A further potential instability arises from units with low coefficients of variation that cannot be replaced by their means due to a strong response to small perturbations.

\textbf{Single-class sensitivity analysis is possible with softmax outputs}. The study findings support the concept of determining single-class sensitivity, which may potentially be utilized to suppress the impact of input distortions on classifiers or other decision-making systems that rely on class probabilities. The proposed approach is based on candidate filtering by the co-occurrence of sensitive classes in predictions obtained through activation damaging. In general, this is not the only way to gather evidence supporting the correctness of extrapolating single-class sensitivity to the entire network from the classification layer. Identifying the classes that are systematically inhibited by the network with edited activations is no less important, as negative feedback and inhibition are inherent in biological neural networks, and this can be implemented by ranking the classifier's confidence in its predictions.

\textbf{Different sensitivity values highlight complementary trends}. Although visual inspection can assist in pattern recognition, any analysis should be performed with caution due to the approximate nature of the results. Complementary analysis was conducted employing cross-validated
LDA, which treats sensitivity maps as features to predict the SA variable, and unit-wise Spearman correlation, which captures monotonic relationships between variables. Both techniques enabled us to reveal the consistent common data patterns. 
Computed sensitivity values were further assessed using guided neural damaging of the corresponding checkpoints. The correlational patterns, as well as single-class sensitivity analysis relating the most sensitive classes to the top-5 masked predictions, both demonstrated specificity across various augmentations and networks.
Certain patterns are consistently repeated for a fixed architecture across both datasets. Given that the work excludes fine-tuned networks pretrained on other datasets (i.e., transfer learned models), it suggests that sensitivity may also be related to the intrinsic properties of an architecture. The resemblance in patterns between similarly structured models (AlexNet and VGG11, ILSVRC) also potentially indicates this point.

\textbf{Distributed nature of specialization}. Exploration of neural networks using the constructed framework has established that, although a set of augmentations affects their elements, different layers and neurons respond inhomogeneously to specific SA variables. Such preferences reflect the specialization of network elements that emerged during learning. The specialization changes with network depth: the closer to the output, the smaller the contribution of augmentations and the higher the contribution of the class variable. In addition, activation maps reveal on-off patterns similar to those in biological neural networks, which also differ depending on the SA variable and the depth of the layer.

Specialization, thus, may be conceived as sensitivity to changes in a certain property of the input data, manifested through the redistribution of the corresponding variable's contribution to the total variance.
The framework proposed in this work is suitable for identifying network components specialized in this sense. In general, these components are not isolated but interfere with each other in terms of Sobol and Shapley sensitivity, as demonstrated by the masked impairement of inference. The top-1 accuracy-based approach used to build the feature space in this analysis provides an initial insight into tendencies, but involves substantial averaging, which became most apparent in the case of ResNet18, as the network most resistant to distortions among convolutional models considered in the current study. To extract correlation patterns for paired sources of distortion (augmentations and masks), it might be worth increasing the feature space dimensionality and using more specific scores, such as per-class accuracy, confusion matrices, and their combinations.

\textbf{Low sensitivity to auxiliary permutation variable.} Introducing the permutation SA variable, which controls the order of applied augmentations, has not resulted in a substantial variance contribution: both the first-order Sobol indices and Shapley values are low for this variable. According to visual inspection and LDA of the corresponding sensitivity maps, weak connections with certain SA variables can be identified, partly due to the small magnitude of the sensitivity values. Therefore, they do not significantly distort the overall results, suggesting that the introduction of the permutation variable is justified.
LDA results for vision transformers exhibit a lower level of contamination by this variable, which may suggest a dependency on the neural architecture and, consequently, the possibility of exploring sensitivity to the augmentation order within the network itself.
In a highly specific scenario, it would be possible to establish a baseline without the variable if the experimental conditions admit a fixed order of augmentations.

\textbf{Isolated network segments depend on both input data and preceding units}.
Analysis of network segments might be relevant for studying large models with expensive inference. It is also important to recognize the relationships between patterns observed in isolated and normal operating modes. By matching these sensitivity patterns, it was confirmed that sensitivity to augmentations depends on both input data and preceding transformations. A comparison of correlation patterns among the standalone units also uncovered that skip connections may facilitate the continuation of sensitivity patterns. That assumption is also supported by Table~{\ref{table:0-accuracy_scores_on_augsets}}, according to which ResNet18 tends to have the smallest percentage decrease in accuracy among CNN models when a single augmentation is applied.
The method utilizes repetition of separate channels from the HSV colour space and may be modified by mapping the values to the expected range of input value changes. Other possible methods for studying isolated segments include generating an activation maximization input to use as a mask for the corresponding channel, or generating multiple inputs of a certain nature, such as fractal patterns, geometric shapes, textures, or noise distributions.

\textbf{Piping inputs to the isolated segments.} The third experimental series is rather an exploratory analysis of the possible connections between different segments of a network. Within the conditions of these experiments, previously computed sensitivity values were used solely for selecting the subset of units. Involving full segments was computationally unreasonable. The main research interest of this part of the work is to understand the possible connections among the segments through their top-performing units. However, there are several difficulties related to the misalignment between the raw input data and the expected inputs. The established approach appears to be poorly suited for connecting the data flow to the cut-off segments, while a medium based on reduced-order modelling may be more promising. However, even the suboptimal piping allowed us to identify several trends suitable for generating further hypotheses regarding data flow assimilation.

\section{Conclusion}
\label{sec:conclusions}

A number of parallels with the functioning of biological neurons were identified within the scope of this work. The latter are typically studied under a treatment-control paradigm involving a single modality~\citep{yang2019study}. At the same time, living systems solve multiple tasks simultaneously and operate in complex environments enriched by multiple multimodal inputs. Having adapted the sampling schemes and set up the SA variables, designed framework appears promising for revealing the specifics of both artificial and biological neural networks under conditions closer to the real world.

It is worth noting that the proposed sensitivity analysis framework utilizes classification specifics only for single-class analysis. Auxiliary variables
incorporate the class and partition selection, as well as the order of augmentations to be applied. In other words, they do not rely on the network structure, making it possible to reuse the framework for networks designed to solve other problems.

On the basis of sensitivity analysis, methods for improving resistance to input distortions, including adversarial attacks, can potentially be developed in the form of guided activation corrections or fine-tuning. A fundamental issue with variance-based methods is the time-consuming necessity to generate a sufficient number of samples. Therefore, real-world applications require online estimation methods, while the developed framework could be utilized to generate warm-start priors.

\section*{Supplementary material}
\label{sec:sup-mat}

Supplementary material associated with this article can be found as arXiv Ancillary files (Supplementary S1, attached) and in the corresponding Zenodo repositories (Supplementary S2--S3; see \hyperref[sec:data-avail]{Data availablility}).

\section*{Data availability}
\label{sec:data-avail}

The present study employs the training and validation partitions of the
ImageNet Large-Scale Visual Recognition Challenge (ILSVRC) dataset, which can be downloaded from the corresponding competition page at Kaggle: \url{https://www.kaggle.com/competitions/imagenet-object-localization-challenge}

ILSVRC uses the subset of the ImageNet dataset as its training partition and provides its own validation (labelled) and test (labels not publicly available) sets.
Since the training part of the ILSVRC dataset is non-uniformly distributed among classes, and the minimum number of samples per class is $732$, the training partition was limited to a subset of $732 \cdot 1000$ samples, controlled by a random seed. The validation partition has already been equalized by class, containing $50 \cdot 1000$ samples.

Another dataset used in the study is Places365~\citep{zhou2017places}, selected for its scale and diversity ($365$ classes), 
acceptable image resolution ($256 \times 256$ version), and the availability of CNN models trained from scratch on this dataset. Similar to ILSVRC, its validation partition has already been equalized ($100$ samples per class), and the training partition was separately balanced to contain $3068$ images per class. Link to the dataset: \url{http://places2.csail.mit.edu}

Computed sensitivity values and related computational experiment results are available at the corresponding Zenodo repositories:\\
\href{https://dx.doi.org/10.5281/zenodo.10499818}{10.5281/zenodo.10499818} (ILSVRC, CNNs),\\
\href{https://dx.doi.org/10.5281/zenodo.18098132}{10.5281/zenodo.18098132} (Places365, CNNs),\\
\href{https://dx.doi.org/10.5281/zenodo.21284279}{10.5281/zenodo.21284279} (ILSVRC, vision transformers).

\section*{Code availability}
\label{sec:code-avail}

All models and computational experiments were implemented in the Python programming
language using the conda package manager (\href{https://dx.doi.org/10.5281/zenodo.10055409}{10.5281/zenodo.10055409}). 
This study employed the following packages:
pytorch~\cite{NEURIPS2019_9015},
SALib~\cite{Herman2017},
h5py~\cite{collette_python_hdf5_2014},
opencv~\cite{opencv_library},
numpy~\cite{harris2020array},
scipy~\cite{2020SciPy-NMeth},
pandas~\cite{mckinney-proc-scipy-2010,reback2020pandas},
scikit-learn~\cite{scikit-learn},
seaborn~\cite{Waskom2021},
matplotlib~\cite{hunter2007matplotlib},
and statsmodels~\cite{seabold2010statsmodels}.
The experiments were systematized using Jupyter Notebooks~\cite{Kluyver2016jupyter}.

All source code, scripts, and Jupyter notebooks produced within this
study can be found at the GitHub repository: \url{https://github.com/kharyuk/activation_sa}

\section*{Acknowledgements}
\label{sec:acknow}

Part of the work related to sensitivity analysis of the inner activations and its validation by guided masking prediction was supported by the Ministry of Science and Higher Education 
grant No. 075-15-2020-801. Single-class sensitivity analysis and sensitivity analysis of the network segments were supported by Ministry of Science and Higher Education grant No. 075-10-2021-068.

\section*{CRediT author statement}
\label{sec:author-contribution}
\textbf{P.K.}: Conceptualization, Methodology, Software, Validation, Investigation, Data curation, Visualization, Writing - Original draft preparation, Reviewing and Editing;
\textbf{S.M.}: Conceptualization, Resources, Writing - Reviewing and Editing;
\textbf{I.O.}: Conceptualization, Resources, Funding acquisition.

S.M. and I.O. proposed the idea to analyze the sensitivity of neuronal activations using Sobol indices.
P.K. 
suggested studying sensitivity to simultaneously applied augmentations, conducted research and developed the idea into the proposed framework.
P.K. and S.M. contributed to the work on the text.
I.O. provided administrative support.

\appendix

\section{Experimental setup details}
\label{appendix:exp-setup}
\setcounter{table}{0}

The section provides details of the internal augmentation parameters (Table~{\ref{table:A0-details-aug-setup}}) and lists number of activations at checkpoints and the number of samples used for computing sensitivity values in the first and third experimental series (Tables~{\ref{table:A1-details-setup1}},~\ref{table:A2-details-setup2}). As mentioned in Section~{\ref{subsect:sobol-indices}}, internal parameters were combined in order to treat augmentations as SA variables.

The number of samples used to compute the rank-biserial correlation coefficients (see remark in S1.2) is calculated as follows: $(1 + n_{\text{aug}} + m_{\text{inner}} \cdot n_{\text{aug}, \boldsymbol{\theta}}) \cdot k \cdot N_{\text{classes}}$, where $n_{\text{aug}, \boldsymbol{\theta}}$ and $n_{\text{aug}}$ are the numbers of augmentations with and without internal parameters, $m_{\text{inner}}$ is the number of samples for parameterized augmentations, $N_{\text{classes}}$ is the number of classes in the dataset, and $k$ is the number of samples per class.

In the case of the sampling scheme 1 (which uses the Salteli sampler), the final number of samples is calculated as $n_{\text{samples}} \cdot (2 \cdot d_i + 2)$, where $n_{\text{samples}}$ is the specified number of samples, which must be a power of 2, and $d_i$ is the number of variables for the $i$th augmentation set. 

The corresponding formula for the case of Shapley values (sampling scheme 2) is  $n_{\text{perm.}} \cdot d_i \cdot n_{\text{outer}} \cdot n_{\text{inner}}$, where $n_{\text{perm.}}$ is the number of permutations, $d_i$ is, as before, the number of variables for the $i$th augmentation set, and $n_{\text{outer}}$ and $n_{\text{inner}}$ are the numbers of outer and inner samples.

\begin{table}[h!]
\centering
\caption{Inner parameters $\boldsymbol{\theta}_i$ of augmentations considered in this study. $\text{Uniform}_{\text{C}}(a, b)$: continuous uniform distribution; $\text{Uniform}_{\text{D}}(a, b)$: discrete uniform distribution. The last column (``related parameters'') lists the values of non-random (constant) parameters. Square brackets $[ \cdot ]$ denote rounding to the nearest integer.}
\label{table:A0-details-aug-setup}
\small
\input{./tables/A0_aug_details.tex}
\end{table}

\begin{table}[h!]
\centering
\caption{Details of the first experimental series: number of activations (listed for each checkpoint and total) and number of samples per unit for each sensitivity value. Rank biserial correlation coefficient (\textit{rbscc}); sampling scheme 1: means and variances (\textit{sitv}), Sobol indices (\textit{si}); sampling scheme 2: means and variances (\textit{shptv}), Shapley values (\textit{shpv}).}
\label{table:A1-details-setup1}
\hspace*{-0.25 cm}
\small
\input{./tables/A1_exp1_details.tex}

\vspace{-0.25 cm}
\end{table}

\begin{table}[tp!]
\centering
\caption{Details of the third experimental series: number of activations (listed for each checkpoint and total) and number of samples per unit for each sensitivity value. Sampling scheme 1: means and variances (\textit{sitv}), Sobol indices (\textit{si}); sampling scheme 2: Shapley values (\textit{shpv}).}
\label{table:A2-details-setup2}
\small
\input{./tables/A2_exp3_details.tex}
\end{table}

\section{Linear discriminant analysis (LDA) of the convolutional sensitivity maps}
\label{appendix:lda}

Predictions of the SA variables
share several common trends across all CNNs. Overall accuracies for the (VGG11, ILSVRC) and (AlexNet, Places365) pairs tend to be correspondingly higher and lower, forming a range within which the accuracy curves (plotted as values at sequential checkpoints) of the other convolutional models are located. Curves of the total Sobol indices tend to be grouped according to the underlying model architecture; the first-order Sobol indices are less prominent in this regard, and the Shapley values mix the tendencies observed for AlexNet and ResNet18.
As checkpoints progress deeper, the overall accuracies tend to
decrease, with certain checkpoints breaking this trend: for AlexNet, features.9; for VGG11, features.10, and features.15; for ResNet18, layer3.
Shapley values preserve higher overall accuracy values, whereas accuracy decreases more rapidly for both Sobol indices.

Single-variable accuracy patterns are similar for all convolutional networks. Erasing, rolling, and class variables tend to be the most separable (high accuracy) on the first augmentation set. On the second augmentation set, the most discriminative variables are horizontal flip, rotation, class, and partition for Shapley values, and elliptic local blur and class for both Sobol indices.
Remarkably, networks tend to be grouped by both their architecture and the dataset in terms of the mean square values of single-variable accuracies. Even AlexNet trained on Places365, which contains many inactive neurons, is closer to AlexNet trained on ILSVRC than to other networks (Sobol indices), or to ResNet18 trained on the Places365 dataset (Shapley values). The (AlexNet, ILSVRC) pair best matches (ResNet18, ILSVRC) for Shapley values and (AlexNet, Places365) for total Sobol indices. ResNet18 tends to align more with its own architecture on both datasets. VGG11 is intermediate---its patterns, as measured by Shapley values, appear closer to those of AlexNet on the same dataset, while in terms of Sobol indices, they are closer to those of ResNet18.

All convolutional networks show similar dependencies on confounding with other variables in terms of Shapley values. Class and partition are mutually entangled; sharpening, grayscaling, and Gaussian blurring, as well as adjusting brightness, contrast, and saturation, tend to be predicted instead of permutation. The permutation itself notably contaminates the predictions of sharpness and elliptic local blur variables, as well as colour transforms. In general, permutation is confused with almost all augmentations except: rolling, where class and partition are more prominent; horizontal flip, which is slightly mispredicted as hue; and rotation, where predictions of the horizontal flip and hue also occur. Colour augmentations from the second augmentation set tend to be mutually mispredicted.
In addition to permutation, erasing predictions lose scores due to Gaussian blur, grayscale, and sharpness variables; grayscale predictions---due to sharpness, and Gaussian blur; Gaussian blur predictions---due to grayscale, sharpness, and partially erasing.

Sobol indices-based predictions produce more uniform confusion values at earlier checkpoints, preventing specific persistent confounding pairs from being identified. Nevertheless, similar patterns to those observed for Shapley values can still be partially traced at certain convolutional checkpoints.

Vision transformers have their own specifics. For ViT-b-16, the CLS features were removed as being inconsistent with the patch-based ones. Its Shapley values are highly separable at all checkpoints. Sobol indices are also appreciably separable and share the separability trend for erasing and elliptic local blur, though with no dependency on the depth.

Swin-t demonstrated the same patterns for Shapley values, preserving their separability at all checkpoints. Its class variable predictions by the first-order Sobol indices approach 1 with increasing depth, while the other variables tend to lose predictability. Erasing, elliptic local blur, and rolling are moderately separable: they retain separability to a greater extent than other variables, but still decrease at several checkpoints. Total Sobol indices hold an intermediate position between first-order Sobol indices and Shapley values.

\section{Hierarchical clustering analysis (HCA) of the damaged predictions}
\label{appendix:hca}

Activations masked during inference change the data flow and the corresponding predictions. The resulting accuracy changes generate the prediction patterns used as a feature space for further correlation analysis. Computed correlation matrices can be transformed into distance matrices by applying the following transform: $\tau(\rho) = \sqrt{1 - |\rho|}$, where $\rho$ is a correlation coefficient. On this basis, hierarchical clustering has been additionally performed using the hierarchical agglomerative clustering algorithm with average linkage (Figures~{S1.6.1 (a)--(c), S1.6.2 (a)--(b)}). HCA allows one to highlight patterns that are already identified through direct inspection of the correlations.

In the case of total Sobol indices, the initial distances between single-sample clusters are minimal on the ILSVRC dataset, making it difficult to match many of them during the first few merges. Thus, the further analysis was limited to first-order Sobol indices and Shapley values.
Pairs with the same masking SA variable tend to group together (especially for ResNet18), which may be explained by the greater impact of guided damaging on the inference compared to input augmentations.

Well-matched augmentations and masks of the same variable from the first augmentation set $A_1$ (Gaussian blur, grayscale, and erasing for AlexNet; erasing, grayscale, and rolling for VGG11) remain as single-element clusters or clusters with a small number of elements until large-scale merging occurs.
For the second augmentation set $A_2$, all networks tend to form early clusters with colour adjustment pairs. These maskings tend to be grouped but are entangled in a non-uniform way, so there are no exact subtrees among either the networks or the sensitivity values.

\section{Threshold selection for single-class sensitivity analysis}
\label{appendix:thresh_scsa}

To isolate the mean Jaccard values for further analysis, one may search for their statistical outliers. The corresponding probabilistic model can be identified as one related to the problem of sampling $n$ balls from two identical boxes with $N$ unique balls in each. The single Jaccard index corresponds to $f(k) = k/(2n-k)$ as a function of $k$, that is, the size of the intersection of two $n$-ball samples, distributed as
\begin{equation}
\label{eq:threshold_scsa}
    p(k) = \frac{\binom{n}{k} \binom{N-n}{n-k} }{ \binom{N}{n}}, \quad k =  \overline{0, n}.
\end{equation}

In other words, $k$ has a hypergeometric distribution with parameters $(N, n, n)$, with $N$ equal to the total number of classes, and $n$ - the number of the most probable classes in a single prediction (top-n). The average value is computed over $s \cdot N$ samples, where $s$ is the number of samples per class. The Monte Carlo method of $10^7$ trials provides the following estimates: (1) $N = 1000$, $s=50$, $n=5$: $q_1 \approx 0.00273$, $q_3 \approx 0.00284$; (2) $N = 365$, $s=100$, $n=5$: $q_1 \approx 0.00755$, $q_3 \approx 0.00775$. Thus, the upper thresholds based on the inter-quantile range, $\tau = q_3 + 3/2\cdot(q_3-q_1)$, are: $\tau \approx 0.003$ for the ILSVRC dataset, and $\tau \approx 0.0081$ for Places365.

The computed thresholds were used to highlight sufficiently high mean Jaccard indices listed in Tables~{S1.5.1, S1.5.2}. The corresponding outlier detection results are also summarized in Tables~{S1.7 (d)--(e)}.

\section{Detailed single-class sensitivity analysis}
\label{appendix:scsa}

The current section presents the single-class sensitivity analysis based on the class-sensitivity levels of classification layers (Tables~{S1.7 (a)--(c)}), with respect to the statistically confirmed outliers listed in Tables~{S1.7 (d)--(e)}.  
This analysis was conducted as follows. In accordance with the specifics of each augmentation, several expectations were formulated for each augmentation SA variable. For each (network, dataset, sensitivity value) triplet, the corresponding sensitive classes confirmed by prediction bias were selected. Images of these classes were visually inspected to explore possible tendencies affecting the selection results. The observations were systematized by augmentation and are summarized in the text below.

Augmentation SA variables selected for this study can be grouped in the following way:
colour adjustments (hue, contrast, saturation, brightness, grayscale), local area distortions (erasing, elliptic local blur), global area distortions (Gaussian blurring, sharpening), and motions (rolling, horizontal flip, rotation).

For classes sensitive to motion augmentations, it is natural to expect either vertical line (horizontal flip) or circular (rotation) symmetry, as well as
vertical and horizontal line patterns (rolling).
In the case of local area distortions, the sensitive classes are expected to have
discriminative features such as
elliptic forms in the frame centre (elliptic local blur) or filled rectangular areas (erasing). As a global transform, sharpening eliminates blurring, sometimes introducing a sand texture pattern and emphasizing contours and shapes, potentially the most important features for predicting classes sensitive in this sense. Gaussian blurring has the opposite effect on images.

When matching class sensitivity with colour adjustments,
it is crucial to recognize how they affect the input,
which can be visualized by changes in colour histograms (channel-wise distributions of intensity values).
By adjusting the brightness, one is moving the histogram as a whole towards a left or right edge.
The hue adjustment shifts histograms relative to each other, changing the dominant colour of the entire image. The contrast adjustment increases (decreases) their variances, highlighting (suppressing) the borders between areas of different colours. The saturation parameter controls the image colourfulness by pulling the histograms apart or pushing them together. The limiting case for the latter is a grayscale transform, where all histograms coincide in the shared area.

Let us start the analysis with the motion augmentations.
Several classes identified as sensitive to the rolling transform barely meet the expectations: for (AlexNet, ILSVRC), \textit{indigo bird}, and \textit{soft-coated wheaten terrier}; for (ResNet18, ILSVRC), \textit{mashed potato}, and \textit{chocolate sauce}. A clue to comprehend the matter resides within the subset of samples that includes the framed pictures, either artificially generated or naturally induced---e.g., for (AlexNet, Sobol, Places365), \textit{throne room}, and \textit{galley}. The observation is consistent across all networks and datasets, although the pattern is not limited to these selections, indicating that the sensitivity ranking depends not only on a dataset but also on a concrete learning outcome.

High prediction bias supporting the class sensitivity to rotation was detected for all considered network-dataset pairs.
As expected, classes with object or frame circular symmetry were encountered, in (ILSVRC): \textit{mushroom}, \textit{hotpot}, \textit{facepowder}, partially---\textit{paddlewheel}, \textit{tray}, and \textit{bathtub}; but most samples of the identified classes are frames taken with a tilted camera or objects captured from different angles, such as, in (ILSVRC), \textit{shoji}, \textit{abacus}, \textit{buckle}, \textit{bathtub}, and \textit{cellular phone}; in (Places365), \textit{office}, and \textit{home office}; as well as photographs of animals in various poses, in (ILSVRC), \textit{coyote}, \textit{hyena}, \textit{weasel}, and \textit{platypus}.

For the horizontal flip, the following common Sobol-sensitive ILSVRC classes
were top-scored by AlexNet and VGG11: \textit{snowmobile}, \textit{shield}, and \textit{missile}. For the latter two classes, a vertically elongated pattern can be traced, as well as for several Places365 classes (\textit{basement}, \textit{staircase}, \textit{forest road}, \textit{general store indoor}, and \textit{trench}). The \textit{cliff dwelling} (ILSVRC) objects are aligned with the natural landscape and can hardly be distinguished from their flipped copies. The same reasoning partially applies to other ILSVRC classes (\textit{snowmobile}, \textit{giant schnauzer}, \textit{moped}, and \textit{Siberian husky}), whose objects are captured mainly from the side.
From another perspective, these latter classes are rather asymmetrical, similar to several sensitive Places365 classes
(\textit{japanese garden}, \textit{hospital room}, \textit{cottage}, \textit{market outdoor}),
allowing us to assume
that horizontal flip sensitivity is tied to frame symmetry. Since there are two types of individual sensitivity maps for horizontal flip---those sensitive either to the symmetry axis or to the remaining field of view---the final-layer sensitivity may relate to both symmetrical and asymmetrical patterns. That is, the classes are ranked with respect not only to their inherent static
properties but also to a kind of invariance under the applied transform and corresponding violations of this invariance.

The next group of augmentations refers to local distortions. The most notable result is the erasing-sensitivity for AlexNet on both datasets
(Sobol, Figure~{\ref{fig:5-sobol-indices-jaccard}}), where the corresponding objects are rectangular screens and monitors (in ILSVRC, \textit{screen}, \textit{monitor}, and \textit{television}; in Places365, \textit{home theater}, \textit{movie theater indoor}, and \textit{television room}) or rectangular-shaped ones (in ILSVRC, \textit{lighter}, and \textit{pedestal}; in Places365, \textit{phone booth}, and \textit{barndoor}). Similar findings are valid for the ResNet18: Sobol indices top-ranked the same \textit{screen}, \textit{monitor}, and \textit{television} (ILSVRC) and \textit{home theater}, \textit{movie theater indoor}, and \textit{television room} (Places365) classes, along with rectangular-shaped \textit{wardrobe} (ILSVRC) and \textit{barndoor} (Places365).
At the same time, it is hard to consistently interpret the classes highlighted by Shapley values for AlexNet and VGG11, as most of them lack filled rectangular patterns. Probably the more abstract pattern of any large filled monochrome or low-coloured areas partially blocking the frame view comes to the forefront.

Elliptic local blur sensitivity can be explained by matching the shape and position of specific discriminative details in the frame. For AlexNet, in (ILSVRC, Sobol):
\textit{golf cart}, and \textit{sports car} objects have round lights or wheels often centred in the frame; \textit{cup} images at several angles appear to contain an elliptic-like region; \textit{black and gold garden spider} class contains contrasting near-elliptical regions which are likely important for their recognition; in (Places365, Shapley), \textit{galley}, and \textit{kitchen} samples contain rounded metal sinks or kettles; \textit{entrance hall}, many rounded arches; \textit{general store indoor}, circled spots of light on the glass bottles. As for ResNet18, in (ILSVRC, Sobol): \textit{lacewing fly} objects have translucent wings of the similar shape; \textit{cleaver}, and \textit{stingray} are also shaped alike and contain round spots; in (Places365, Sobol), \textit{lecture room}, a subset of samples contains microphones with heads of similar shape; \textit{youth hostel}, lamps with rounded lampshades; \textit{greenhouse indoor}, rounded pots and planters.

Similar reasoning may be applied to sensitive classes of VGG11: in (ILSVRC, Sobol), \textit{barrel (cask)}, several objects either have a circular hole in the centre or are captured from the top; \textit{parking meter} also contains a round filled pattern; \textit{coyote}, in many pictures the animal is photographed from a distance, occupying a small part in the centre and intersecting with the distortion area;
\textit{cornet}, certain images capture glaring musical instruments facing the bells; in (ILSVRC, Shapley), \textit{scabbard}, many short metal daggers, partially resembling the elliptic shape; \textit{computer mouse}, \textit{mosque} (the main dome), and \textit{wreck}, shapes similar to the distortion. However, it is difficult to follow the same logic to interpret some other classes, e.g., \textit{cliff dwelling}, as sensitive to this augmentation.

But VGG11 is also sensitive to \textit{cliff dwelling} under colour adjustments (for Sobol, contrast, saturation, and hue). This class is represented by pictures of cliffs, canyons, and stone dwellings in the highland areas, and many of these objects are exposed to direct sunlight. Some of these photos have been taken from shade, some from the air. Thus, there is a contrast of light and shade and a prevalence of monochromatic areas (brown tones, occasionally diluted by green grass and blue sky).

Two networks, ResNet18 (Sobol) and VGG11 (Sobol and Shapley), showed the sensitivity of the \textit{jigsaw puzzle} class to the contrast adjustment. The images of this class are characterized by the bright colours of the puzzle picture and the sharp edges of its pieces, and the contrast tuning influences the visibility of these edges.
Similar logic may explain why the following classes were identified as contrast-sensitive:
for AlexNet, in (Sobol, ILSVRC), \textit{walking stick}, \textit{red-breasted merganser}, \textit{pay-phone}, and \textit{chain}; for VGG11, in (Shapley, ILSVRC), \textit{parking meter}, \textit{gyromitra}, and \textit{waffle iron}; for VGG11, in (Sobol, ILSVRC), \textit{cornet} and \textit{shovel}; for ResNet18, in (Sobol, ILSVRC), \textit{bubble}, \textit{spotlight}, and \textit{hook (claw)}; for ResNet18, in (Shapley, ILSVRC), \textit{drumstick}, \textit{wing}, and \textit{holster}.
In addition, sensitivity to contrast may be attributed to lighting conditions or colour palette affecting the boundary visibility of the key scene objects: for AlexNet, in (Sobol, Places365), \textit{television studio}, \textit{music studio}, \textit{childs room}, \textit{amusement arcade}, and \textit{stage indoor};  for ResNet18, in (Sobol, Places365), \textit{desert vegetation}, \textit{television studio}, and \textit{kindergarden classroom}; for AlexNet, in (Shapley, ILSVRC), \textit{napkin}, \textit{pig}, and \textit{African crocodile}.

Unlike the previous augmentation, saturation affects the spacing between channel-wise pixel intensity histograms.
The corresponding sensitive classes were identified for almost all networks except ResNet18 (Places365, Shapley), and share the following basic patterns: there are traces of image processing visible in several samples (e.g., \textit{hay}, \textit{bubble}); respective objects contain long monochromatic patches (\textit{refrigerator}, \textit{upright piano}, \textit{screen}, \textit{water bottle}, and \textit{dough}) or distinctive monochrome elements (\textit{holster}, \textit{wardrobe}, \textit{archaeological excavation}, \textit{trench}, and \textit{rotisserie}); respective objects are set against large monochrome elements of the scene (\textit{barbell}, \textit{sea urchin}); monochrome illumination of the scene or dominating palette of limited colours (\textit{discotheque}, \textit{ice shelf}, \textit{stage indoor}, \textit{ice floe}, and \textit{wet bar}).

Importantly, the classes sensitive to grayscale transformation differ from saturation-sensitive ones. One of them, the \textit{tree frog} class (Sobol, ILSVRC), is listed for all three CNNs. The corresponding images are mainly in green tones, suggesting that the inference is non-reliant on the colour feature. The latter characteristic appears essential for several other similar classes: for (VGG11, Sobol, ILSVRC), \textit{warthog}, \textit{scuba diver}, and \textit{Russian wolfhound}; for (ResNet18, Sobol, ILSVRC), \textit{chain}; for (ResNet18, Shapley, ILSVRC), \textit{lakeside}; for (AlexNet, Shapley, Places365), \textit{mountain path}, \textit{lawn}, and \textit{sandbox}. Similarly, the shape of object may be more important for predicting several classes, including: for (ResNet18, Sobol, ILSVRC), \textit{sweatshirt} and \textit{lifeboat}; for (ResNet18, Shapley, ILSVRC), \textit{coral reef} and \textit{wig}. Another possible factor is the presence of grayscale and sepia images among the samples: for (AlexNet, Sobol, Places365), \textit{hospital room}, \textit{archive}, \textit{music studio}, \textit{coffee shop}, and \textit{physics laboratory}.

As for hue-sensitive classes, two major patterns related to the dominant colour and palette changes can be observed. Samples of the first type contain the dominant background colour with shades varying, including variability caused by artificial illumination or differences in natural light level (e.g., in ILSVRC, \textit{agaric}, \textit{missile}, \textit{Chihuahua}, \textit{house finch}, \textit{salamandra}, \textit{breastplate}, and \textit{bee}; in Places365, \textit{stage indoor}, \textit{rainforest}, \textit{ruin}, \textit{beer hall}, \textit{plaza}, and \textit{discotheque}).
The second recognizable pattern emphasizes the colour of the central object or its critical feature, especially for close-up shots (in ILSVRC, \textit{digital clock}; in Places365, \textit{botanical garden}, \textit{vegetable garden}, and \textit{florist shop indoor}). Several hue-sensitive classes combine both patterns (in ILSVRC, \textit{slum}, \textit{jeep}, and \textit{jellyfish}; in Places365, \textit{recreation room}).

Similarly, one may isolate the following pair of typical patterns for brightness-sensitive classes:
rich, colourful content (in ILSVRC, \textit{grey wolf}, \textit{howler monkey}, and \textit{bookshop}; and in Places365, \textit{childs room}, \textit{playroom}, \textit{candy store}, \textit{art studio}, \textit{pizzeria}, and \textit{bazaar indoor}); various luminosity levels, often dependent on lighting (in ILSVRC, \textit{worm fence}, \textit{boxer}, and \textit{cliff (drop)}; and in Places365, \textit{creek}, \textit{patio}, \textit{crosswalk}, \textit{rock arch}, and \textit{corral}); and a combination of both (in ILSVRC, \textit{pinwheel}, \textit{plate}, and \textit{king snake}; and in Places365, \textit{general store outdoor}).

The latter group of augmentations consists of Gaussian blur and sharpening, both of which impact the same high-level image feature---sharpness. Gaussian blur cuts off fine details, complicating the class discrimination. Such fine-detailed classes sensitive to blurring are: in Places365, \textit{throne room}, \textit{temple asia}, and \textit{server room}. Similarly, several other classes include low resolution samples: in ILSVRC, \textit{broom} and \textit{hair slide}; and in Places365, \textit{restaurant}, \textit{market indoor}, \textit{children room}, \textit{shoe shop}, \textit{kitchen}, \textit{coffee shop}, and \textit{playroom}. Others contain defocused or blurred samples: in ILSVRC, \textit{scuba diver}, \textit{bubble}, and \textit{green snake}. Finally, its sensitivity may be explained by the inherent coarseness of the objects: in ILSVRC, \textit{mask}, \textit{coho salmon}, \textit{flat-coated retriever}, \textit{affenpinscher}, and \textit{dingo}; and in Places365, \textit{shed}, \textit{barndoor}, \textit{driveway}, \textit{basement}, and \textit{fabric store}.

Sharpness adjustment may introduce a sandy texture or amplify grain artifacts. Several classes identified as sensitive to this augmentation exhibit similar patterns: in ILSVRC, \textit{warthog}, \textit{coho salmon}, \textit{Alaskan malamute}, \textit{poncho}, \textit{harvester  (reaper)}, and \textit{face powder}; and in Places365, \textit{excavation} and \textit{lawn}. Similar to those sensitive to blurring, there are also classes that contain fine details. Images of these classes tend to be diverse and colourful, and are prone to confusion with other classes where similar details emerge due to augmentation artifacts: in ILSVRC, \textit{church}; and in Places365, \textit{delicatessen}, \textit{bazaar outdoor}, \textit{church outdoor}, and \textit{temple asia}. Other sensitive classes feature various ripples and edges that are emphasized by the transformation: in ILSVRC, \textit{banister}, \textit{bee eater}, \textit{leaf beetle}, \textit{overskirt}, \textit{carton}, \textit{Windsor tie}, and \textit{grasshopper}; in Places365, \textit{harbor} and \textit{elevator lobby}.


\input{./article.bbl}
\bibliographystyle{elsarticle-num-names}

\end{document}

%% file: front_matter.tex
\title{Exploring specialization and sensitivity of convolutional neural networks under simultaneous image augmentations}

\author[1,2]{Pavel~Kharyuk\corref{cor1}}
\ead{pavel.kharyuk@gmail.com}

\author[3,1]{Sergey~Matveev}
\author[4,2]{Ivan~Oseledets}

\affiliation[1]{
    organization={Marchuk Institute of Numerical Mathematics},
    addressline={Gubkin str., 8},
    postcode={119333},
    city={Moscow},
    country={Russia},
}
\affiliation[2]{
    organization={Skolkovo Institute of Science and Technology},
    addressline={Bolshoy Boulevard, 30, p.1},
    postcode={121205},
    city={Moscow},
    country={Russia},
}
\affiliation[3]{
    organization={Faculty of Computational Mathematics and Cybernetics, Lomonosov Moscow State University},
    addressline={Leninskie gory, 1/52},
    postcode={119991},
    city={Moscow},
    country={Russia},
}
\affiliation[4]{
    organization={Artificial Intelligence Research Institute (AIRI)},
    city={Moscow},
    country={Russia},
}
\cortext[cor1]{Corresponding author}

\begin{abstract}
    Drawing parallels with the way biological networks are studied, the authors adapt the treatment--control paradigm to explainable artificial intelligence research and enrich it through multi-parametric input alterations. A framework for investigating internal inference impacted by input data augmentations is proposed to address the real-world scenario with multiple simultaneously applied treatments instead of observing a single effect. The internal changes in network operation are reflected in activation changes measured by variance, which can be decomposed into components related to each augmentation by employing Sobol indices and Shapley values. To facilitate distinguishing model-related specifics from the self-interactions of augmentations, an auxiliary variable responsible for their permutations has been incorporated into the analysis. The resulting quantities enable one to visualize sensitivity to different variables and use them as features for further analysis and processing, including guided masking of activations. In addition, a method for single-class sensitivity analysis is introduced through a pipeline that involves class selection according to their match with the prediction bias generated by targeted damaging of the activations. Three convolutional networks, namely AlexNet, VGG11, and ResNet18, have been evaluated via the designed framework, with two vision transformers, ViT-b-16 and Swin-t, partially complementing the findings. Two sets of augmentations, consisting of five and seven distinct input transformations applied simultaneously, have been utilized as sources of variance in the image domain. Relying on the observed parallels, the authors assume that the developed framework can potentially be transferred to the study of biological neural networks in complex environments, as well as enhance the design of autonomous decision-making systems.
\end{abstract}

\begin{keyword}
Explainable AI \sep
Shapley values \sep
Sobol indices \sep
Convolutional neural networks \sep
Sensitivity analysis \sep
Neural network specialization
\end{keyword}

%% file: tables/0_accuracy_scores_part_A.tex
\begin{tabular}{
    l@{\hskip 0.15cm}
    l@{\hskip 0.15cm}
    c@{\hskip 0.2cm}
    c@{\hskip 0.2cm}
    c@{\hskip 0.2cm}
    c@{\hskip 0.2cm}
    c@{\hskip 0.2cm}
    c@{\hskip 0.2cm}
}
\hline
 \textbf{} & \textbf{} & \textbf{Orig.} & \textbf{Erasing} & \textbf{Sharp.} & \textbf{Rolling} & \textbf{Grayscale} & \textbf{Gaus.blur} \\
\hline
\multicolumn{8}{c}{\textit{Imagenet (ILSVRC), CNNs}}\\
\textbf{AlexNet} & \textit{top-1} & \text{0.5631} & \text{\underline{0.3969}} & \text{0.5588} & \text{0.5036} & \text{\underline{0.3395}} & \text{0.4511} \\
\text{} & \textit{top-5} & \text{0.7895} & \text{\underline{0.6269}} & \text{0.7866} & \text{0.7413} & \text{\underline{0.5767}} & \text{0.6927} \\
\textbf{VGG11} & \textit{top-1} & \text{0.6888} & \text{\underline{0.5874}} & \text{0.6862} & \text{0.6519} & \text{\underline{0.5497}} & \text{0.5802} \\
\text{} & \textit{top-5} & \text{0.8854} & \text{\underline{0.8106}} & \text{0.8843} & \text{0.862\hphantom{0}} & \text{\underline{0.7926}} & \text{0.8102} \\
\textbf{ResNet18} & \textit{top-1} & \text{0.6967} & \text{\underline{0.5707}} & \text{0.6937} & \text{0.6594} & \text{\underline{0.5589}} & \text{0.6187} \\
\text{} & \textit{top-5} & \text{0.89\hphantom{00}} & \text{\underline{0.7932}} & \text{0.8901} & \text{0.8657} & \text{\underline{0.8004}} & \text{0.8363} \\
\multicolumn{8}{c}{\textit{Imagenet (ILSVRC), vision transformers}}\\
\textbf{ViT-b-16} & \textit{top-1} & \text{0.8098} & \text{0.7852} & \text{0.8084} & \text{0.7947} & \text{\underline{0.7131}} & \text{\underline{0.764}\hphantom{0}} \\
\text{} & \textit{top-5} & \text{0.9527} & \text{0.9388} & \text{0.9521} & \text{0.9433} & \text{\underline{0.9022}} & \text{\underline{0.9294}} \\
\textbf{Swin-t} & \textit{top-1} & \text{0.8105} & \text{0.7906} & \text{0.8123} & \text{0.798\hphantom{0}} & \text{\underline{0.7565}} & \text{\underline{0.7318}} \\
\text{} & \textit{top-5} & \text{0.956\hphantom{0}} & \text{0.9457} & \text{0.9558} & \text{0.9488} & \text{\underline{0.9289}} & \text{\underline{0.9142}} \\
\textbf{MaxViT-t} & \textit{top-1} & \text{0.8337} & \text{0.8113} & \text{0.8353} & \text{0.8204} & \text{\underline{0.7915}} & \text{\underline{0.7846}} \\
\text{} & \textit{top-5} & \text{0.9662} & \text{0.9547} & \text{0.9667} & \text{0.9588} & \text{\underline{0.9478}} & \text{\underline{0.9438}} \\
\multicolumn{8}{c}{\textit{Places365, CNNs}}\\
\textbf{AlexNet} & \textit{top-1} & \text{0.473\hphantom{0}} & \text{\underline{0.3288}} & \text{0.4619} & \text{0.3934} & \text{\underline{0.2856}} & \text{0.4056} \\
\text{} & \textit{top-5} & \text{0.7776} & \text{\underline{0.6179}} & \text{0.7698} & \text{0.7001} & \text{\underline{0.5594}} & \text{0.7081} \\
\textbf{ResNet18} & \textit{top-1} & \text{0.535\hphantom{0}} & \text{\underline{0.4411}} & \text{0.5235} & \text{0.494\hphantom{0}} & \text{\underline{0.4209}} & \text{0.4783} \\
\text{} & \textit{top-5} & \text{0.8363} & \text{\underline{0.7544}} & \text{0.8274} & \text{0.8015} & \text{\underline{0.729\hphantom{0}}} & \text{0.7894} \\
\hline
\end{tabular}

%% file: tables/0_accuracy_scores_part_B.tex
\begin{tabular}{
    l@{\hskip 0.15cm}
    l@{\hskip 0.15cm}
    c@{\hskip 0.2cm}
    c@{\hskip 0.2cm}
    c@{\hskip 0.2cm}
    c@{\hskip 0.2cm}
    c@{\hskip 0.2cm}
    c@{\hskip 0.2cm}
    c@{\hskip 0.2cm}
    c@{\hskip 0.2cm}
}
\hline
\textbf{} & \textbf{} & \textbf{Orig.} & \textbf{Bright.} & \textbf{Contrast.} & \textbf{Satur.} & \textbf{Hue} & \textbf{Hflip} & \textbf{Rotation} & \textbf{El.loc.blur} \\
\hline
\multicolumn{10}{c}{\textit{Imagenet (ILSVRC), CNNs}}\\
\textbf{AlexNet} & \textit{top-1} & \text{0.5631} & \text{0.496\hphantom{0}} & \text{\underline{0.4627}} & \text{\underline{0.47}\hphantom{00}} & \text{\underline{0.3392}} & \text{0.5622} & \text{0.491\hphantom{0}} & \text{0.5499} \\
\text{} & \textit{top-5} & \text{0.7895} & \text{0.7265} & \text{\underline{0.6976}} & \text{\underline{0.7114}} & \text{\underline{0.5649}} & \text{0.79\hphantom{00}} & \text{0.7256} & \text{0.78\hphantom{00}}\\
\textbf{VGG11} & \textit{top-1} & \text{0.6888} & \text{0.6253} & \text{\underline{0.5969}} & \text{0.6323} & \text{\underline{0.5164}} & \text{0.6879} & \text{\underline{0.6034}} & \text{0.6769}\\
\text{} & \textit{top-5} & \text{0.8854} & \text{0.8354} & \text{\underline{0.8184}} & \text{0.8528} & \text{\underline{0.7581}} & \text{0.8856} & \text{\underline{0.8212}} & \text{0.8785}\\
\textbf{ResNet18} & \textit{top-1} & \text{0.6967} & \text{0.6402} & \text{\underline{0.6163}} & \text{0.6537} & \text{\underline{0.5534}} & \text{0.6962} & \text{\underline{0.6257}} & \text{0.684\hphantom{0}}\\
\text{} & \textit{top-5} & \text{0.89\hphantom{00}} & \text{0.8471} & \text{\underline{0.8322}} & \text{0.8658} & \text{\underline{0.7914}} & \text{0.8909} & \text{\underline{0.8371}} & \text{0.8821}\\
\multicolumn{10}{c}{\textit{Imagenet (ILSVRC), vision transformers}}\\
\textbf{ViT-b-16} & \textit{top-1} & \text{0.8098} & \text{0.7858} & \text{\underline{0.7817}} & \text{0.7922} & \text{\underline{0.7029}} & \text{0.8085} & \text{\underline{0.7775}} & \text{0.803\hphantom{0}} \\
\text{} & \textit{top-5} & \text{0.9527} & \text{\underline{0.9382}} & \text{0.9383} & \text{0.9445} & \text{\underline{0.897}\hphantom{0}} & \text{0.9524} & \text{\underline{0.935}\hphantom{0}} & \text{0.9498} \\
\textbf{Swin-t} & \textit{top-1} & \text{0.8105} & \text{0.7941} & \text{\underline{0.7908}} & \text{0.8021} & \text{\underline{0.724}\hphantom{0}} & \text{0.8101} & \text{\underline{0.7633}} & \text{0.8059} \\
\text{} & \textit{top-5} & \text{0.956\hphantom{0}} & \text{0.9467} & \text{\underline{0.9452}} & \text{0.9515} & \text{\underline{0.9119}} & \text{0.956\hphantom{0}} & \text{\underline{0.9286}} & \text{0.954\hphantom{0}} \\
\textbf{MaxViT-t} & \textit{top-1} & \text{0.8337} & \text{0.8208} & \text{\underline{0.8193}} & \text{0.8271} & \text{\underline{0.7721}} & \text{0.8336} & \text{\underline{0.7857}} & \text{0.8286} \\
\text{} & \textit{top-5} & \text{0.9662} & \text{0.9594} & \text{\underline{0.9587}} & \text{0.963\hphantom{0}} & \text{\underline{0.9395}} & \text{0.9663} & \text{\underline{0.9424}} & \text{0.9634} \\
\multicolumn{10}{c}{\textit{Places 365, CNNs}}\\
\textbf{AlexNet} & \textit{top-1} & \text{0.473\hphantom{0}} & \text{0.4122} & \text{\underline{0.3951}} & \text{\underline{0.3969}} & \text{\underline{0.2967}} & \text{0.4729} & \text{0.3998} & \text{0.4676} \\
\text{} & \textit{top-5} & \text{0.7776} & \text{0.7121} & \text{\underline{0.693}\hphantom{0}} & \text{0.7008} & \text{\underline{0.5641}} & \text{0.7784} & \text{\underline{0.6979}} & \text{0.7733} \\
\textbf{ResNet18} & \textit{top-1} & \text{0.535\hphantom{0}} & \text{0.4826} & \text{\underline{0.4581}} & \text{0.4915} & \text{\underline{0.4184}} & \text{0.533\hphantom{0}} & \text{\underline{0.4776}} & \text{0.528\hphantom{0}} \\
\text{} & \textit{top-5} & \text{0.8363} & \text{0.7857} & \text{\underline{0.7641}} & \text{0.8001} & \text{\underline{0.7233}} & \text{0.835\hphantom{0}} & \text{\underline{0.7848}} & \text{0.8312} \\
\hline
\end{tabular}

%% file: tables/A0_aug_details.tex
\begin{tabular}{
    p{1. cm}@{\hskip 0.15cm}
    p{2.975 cm}@{\hskip 0.15cm}
    p{2.55 cm}@{\hskip 0.15cm} 
    p{5.5 cm}@{\hskip 0.15cm} 
    p{3.45 cm}@{\hskip 0.15cm}
}
\hline
{} & {\bf Augmentation} & {\bf Parameter} & {\bf Distribution} & {\bf Related parameters}\\
\hline
{$(A_1.1)$} & {Erasing} &
{$\text{center}_x$}
& {$\text{Uniform}_{\text{D}}(w_{\text{offset}}, \text{width} - w_{\text{offset}})$}
& {$w_{\text{offset}} = [ \text{width} / 5 ]$}\\
{} & {} &
{$\text{center}_y$}
& {$\text{Uniform}_{\text{D}}(h_{\text{offset}}, \text{height} - h_{\text{offset}})$}
& {$h_{\text{offset}} = [ \text{height} / 5 ]$}\\
{} & {} &
{$\text{scale factor}$}
& {$\text{Uniform}_{\text{C}}(0.02, 0.33)$}
& {}\\
{} & {} &
{$\text{aspect ratio}$}
& {$\text{LogUniform}(0.3, 3.3)$}
& {}\\
{$(A_1.2)$} & {Sharpness adj.} &
{--} & {--} & {sharpness factor = 1.5} \\
{$(A_1.3)$} & {Rolling} &
{horizontal shift}
& {$\text{Uniform}_{\text{D}}(-w_{\text{offset}}, w_{\text{offset}})$}
& {$w_{\text{offset}} = [\text{width} / 5]$}\\
{} & {} &
{vertical shift}
& {$\text{Uniform}_{\text{D}}(-h_{\text{offset}}, h_{\text{offset}})$}
& {$h_{\text{offset}} = [\text{height} / 5]$}\\
{$(A_1.4)$} & {Grayscale} &
{--} & {--} & {--} \\
{$(A_1.5)$} & {Gaussian blur} &
{$\sigma_{\text{blur}}$}
& {$\text{Uniform}_{\text{C}}(0.1, 2)$}
& {$\text{kernel size} = (7, 7)$}\\
{$(A_2.1)$} & {Brightness adj.} &
{brightness factor}
& {$\text{Uniform}_{\text{C}}(0.1, 2.5)$}
& {--}\\
{$(A_2.2)$} & {Contrast adj.} &
{contrast factor}
& {$\text{Uniform}_{\text{C}}(0.01, 3.5)$}
& {--}\\
{$(A_2.3)$} & {Saturation adj.} &
{saturation factor}
& {$\text{Uniform}_{\text{C}}(0.01, 3.5)$}
& {--}\\
{$(A_2.4)$} & {Hue adj.} &
{hue factor}
& {$\text{Uniform}_{\text{C}}(-0.5, 0.5)$}
& {--}\\
{$(A_2.5)$} & {Horizontal flip} &
{--} & {--} & {--} \\
{$(A_2.6)$} & {Rotation} &
{angle (degree)}
& {$\text{Uniform}_{\text{C}}(-10, 10)$}
& {--}\\
{$(A_2.7)$} & {Elliptic local blur} &
{angle (degree)}
& {$\text{Uniform}_{\text{C}}(-15, 15)$}
& {$\sigma_{\text{blur}} = 2$}\\
{} & {} &
{semi-major axis}
& {$\text{Uniform}_{\text{D}}(10, 32)$}
& {}\\
{} & {} &
{semi-minor axis}
& {$\text{Uniform}_{\text{D}}(10, 32)$}
& {}\\
{} & {} &
{$\text{center shift}_x$}
& {$\text{Uniform}_{\text{D}}(-32, 32)$}
& {}\\
{} & {} &
{$\text{center shift}_y$}
& {$\text{Uniform}_{\text{D}}(-32, 32)$}
& {}\\
\hline
\end{tabular}

%% file: tables/A1_exp1_details.tex
\begin{tabular}{
    p{0.95 cm}@{\hskip 0.01cm}
    p{0.75cm}@{\hskip 0.15cm}
    p{2.3 cm}@{\hskip 0.15cm} 
    p{2.3 cm}@{\hskip 0.15cm} 
    p{2.3 cm}@{\hskip 0.15cm}
    p{2.3 cm}@{\hskip 0.15cm} 
    p{2.3 cm}@{\hskip 0.01cm}
}
\hline
{} & {} & \textbf{AlexNet} & \textbf{VGG11} & \textbf{ResNet18} & \textbf{ViT-b-16} & \textbf{Swin-t} \\
\hline
  \multicolumn{7}{c}{\textit{\textbf{Number of activations}}}\\
{} & {} & {
\hbox{
$\underbrace{46656}_{\makebox[0pt]{\text{\tiny features.2}}}$
+ $\underbrace{32448}_{\makebox[0pt]{\text{\tiny features.5}}}$
}
\hbox{
+ $\underbrace{64896}_{\makebox[0pt]{\text{\tiny features.7}}}$
+ $\underbrace{3264}_{\makebox[0pt]{\text{\tiny features.9}}}$
}
\hbox{
+ $\underbrace{9216}_{\makebox[0pt]{\text{\tiny avgpool}}}$
+ $\underbrace{4096}_{\makebox[0pt]{\text{\tiny classifier.2}}}$
}
\hbox{
+ $\underbrace{4096}_{\makebox[0pt]{\text{\tiny classifier.5}}}$
+ $\underbrace{1000}_{\makebox[0pt]{\text{\tiny classifier.6}}}$
}
\hbox{= 205672}
} 
& {
\hbox{
$\underbrace{802816}_{\makebox[0pt]{\text{\tiny features.2}}}$
+ $\underbrace{401408}_{\makebox[0pt]{\text{\tiny features.5}}}$
}
\hbox{
+ $\underbrace{200704}_{\makebox[0pt]{\text{\tiny features.10}}}$
+ $\underbrace{100352}_{\makebox[0pt]{\text{\tiny features.15}}}$
}
\hbox{
+ $\underbrace{25088}_{\makebox[0pt]{\text{\tiny avgpool}}}$
+ $\underbrace{4096}_{\makebox[0pt]{\text{\tiny classifier.1}}}$
}
\hbox{
+ $\underbrace{4096}_{\makebox[0pt]{\text{\tiny classifier.4}}}$
+ $\underbrace{1000}_{\makebox[0pt]{\text{\tiny classifier.6}}}$
}
\hbox{= 1539560}
} 
& {
\hbox{
$\underbrace{200704}_{\makebox[0pt]{\text{\tiny maxpool}}}$
+ $\underbrace{200704}_{\makebox[0pt]{\text{\tiny layer1}}}$
}
\hbox{
+ $\underbrace{100352}_{\makebox[0pt]{\text{\tiny layer2}}}$
+ $\underbrace{50176}_{\makebox[0pt]{\text{\tiny layer3}}}$
}
\hbox{
+ $\underbrace{512}_{\makebox[0pt]{\text{\tiny avgpool}}}$
+ $\underbrace{1000}_{\makebox[0pt]{\text{\tiny fc}}}$
}
\hbox{= 553448}
} 
& {
\hbox{
$\underbrace{151296}_{\makebox[0pt]{\text{\tiny enc.layer 2}}}$
+ $\underbrace{151296}_{\makebox[0pt]{\text{\tiny enc.layer 5}}}$
}
\hbox{
+ $\underbrace{151296}_{\makebox[0pt]{\text{\tiny enc.layer 8}}}$
+ $\underbrace{151296}_{\makebox[0pt]{\text{\tiny enc.layer 11}}}$
}
\hbox{
+ $\underbrace{1000}_{\makebox[0pt]{\text{\tiny heads}}}$
}
\hbox{= 606184}
} 
& {
\hbox{
$\underbrace{301056}_{\makebox[0pt]{\text{\tiny features.1}}}$ 
+ $\underbrace{150528}_{\makebox[0pt]{\text{\tiny features.3}}}$ 
}
\hbox{
+ $\underbrace{75264}_{\makebox[0pt]{\text{\tiny features.5}}}$
+ $\underbrace{37632}_{\makebox[0pt]{\text{\tiny features.7}}}$
}
\hbox{
+ $\underbrace{1000}_{\makebox[0pt]{\text{\tiny head}}}$
}
\hbox{= 565480}
} \\ 
\noalign{\vskip -0.3 cm} 
\multicolumn{7}{c}{\textit{\textbf{Samples per activation}}}\\
 {\it rbscc}  & {}
& {\hbox{$(1 + 3 + 9\cdot10) \cdot 50$} \hbox{$\cdot 1000 = 4700000$} }
& {\hbox{$(1 + 3 + 9\cdot10) \cdot 10$} \hbox{$\cdot 1000 = 940000$} }
& {\hbox{$(1 + 3 + 9\cdot10) \cdot 50$} \hbox{$\cdot 1000 = 4700000$} }
& {\hbox{--}}
& {\hbox{--}} \\
 {\it sitv}  & {$(A_1)$}
& {\hbox{$2^{16} \cdot (2\cdot 8 + 2)$} \hbox{$ = 1179648$} }
& {\hbox{$2^{14} \cdot (2\cdot 8 + 2)$} \hbox{$ = 294912$} }
& {\hbox{$2^{16} \cdot (2\cdot 8 + 2)$} \hbox{$ = 1179648$} }
& {\hbox{$2^{15} \cdot (2\cdot 8 + 2)$} \hbox{$ = 589824$} }
& {\hbox{$2^{15} \cdot (2\cdot 8 + 2)$} \hbox{$ = 589824$} }
\\[-0.35cm]
 {}  & {$(A_2)$}
& {\hbox{$2^{16} \cdot (2\cdot 10 + 2)$} \hbox{$ = 1441792$} }
& {\hbox{$2^{14} \cdot (2\cdot 10 + 2)$} \hbox{$ = 360448$} }
& {\hbox{$2^{16} \cdot (2\cdot 10 + 2)$} \hbox{$ = 1441792$} }
& {\hbox{$2^{15} \cdot (2\cdot 10 + 2)$} \hbox{$ = 720896$} }
& {\hbox{$2^{15} \cdot (2\cdot 10 + 2)$} \hbox{$ = 720896$} }
\\[-0.35cm]
 {\it si} & {$(A_1)$}
& {\hbox{$2^{17} \cdot (2\cdot 8 + 2)$} \hbox{$ = 2359296$} }
& {\hbox{$2^{15} \cdot (2\cdot 8 + 2)$} \hbox{$ = 589824$} }
& {\hbox{$2^{17} \cdot (2\cdot 8 + 2)$} \hbox{$ = 2359296$} }
& {\hbox{$2^{16} \cdot (2\cdot 8 + 2)$} \hbox{$ = 1179648$} }
& {\hbox{$2^{16} \cdot (2\cdot 8 + 2)$} \hbox{$ = 1179648$} }
\\[-0.35cm]
 {} & {$(A_2)$}
& {\hbox{$2^{17} \cdot (2\cdot 10 + 2)$} \hbox{$ = 2883584$} }
& {\hbox{$2^{15} \cdot (2\cdot 10 + 2)$} \hbox{$ = 720896$} }
& {\hbox{$2^{17} \cdot (2\cdot 10 + 2)$} \hbox{$ = 2883584$} }
& {\hbox{$2^{16} \cdot (2\cdot 10 + 2)$} \hbox{$ = 1441792$} }
& {\hbox{$2^{16} \cdot (2\cdot 10 + 2)$} \hbox{$ = 1441792$} }
\\[-0.35cm]
 {\it shptv} & {$(A_1)$}
& {\hbox{$2^{18} = 262144$} }
& {\hbox{$2^{18} = 262144$} }
& {\hbox{$2^{18} = 262144$} }
& {\hbox{$2^{18} = 262144$} }
& {\hbox{$2^{18} = 262144$} }
\\[-0.35cm]
  {} & {$(A_2)$}
& {\hbox{$2^{18} = 262144$} }
& {\hbox{$2^{18} = 262144$} }
& {\hbox{$2^{18} = 262144$} }
& {\hbox{$2^{18} = 262144$} }
& {\hbox{$2^{18} = 262144$} }
\\[-0.35cm]
 {\it shpv} & {$(A_1)$}
& {\hbox{$2^{8} \cdot 15 \cdot 2^6 \cdot 2^4$} \hbox{$ = 3932160$} }
& {\hbox{$2^{7} \cdot 15 \cdot 2^5 \cdot 2^3$} \hbox{$ = 491520$} }
& {\hbox{$2^{8} \cdot 15 \cdot 2^6 \cdot 2^3$} \hbox{$ = 1966080$} }
& {\hbox{$2^{8} \cdot 15 \cdot 2^5 \cdot 2^3$} \hbox{$ = 983040$} }
& {\hbox{$2^{8} \cdot 15 \cdot 2^5 \cdot 2^3$} \hbox{$ = 983040$} }
\\[-0.35cm]
 {} & {$(A_2)$}
& {\hbox{$2^{8} \cdot 20 \cdot 2^6 \cdot 2^4$} \hbox{$ = 5242880$} }
& {\hbox{$2^{7} \cdot 20 \cdot 2^5 \cdot 2^3$} \hbox{$ = 655360$} }
& {\hbox{$2^{8} \cdot 20 \cdot 2^6 \cdot 2^3$} \hbox{$ = 2621440$} }
& {\hbox{$2^{8} \cdot 20 \cdot 2^5 \cdot 2^3$} \hbox{$ = 1310720$} }
& {\hbox{$2^{8} \cdot 20 \cdot 2^5 \cdot 2^3$} \hbox{$ = 1310720$} }
\\
\hline
\end{tabular}

%% file: tables/A2_exp3_details.tex
\begin{tabular}{
    p{0 cm}@{\hskip 0.05cm}
    p{1.05 cm}@{\hskip 0.05cm}
    p{0.75cm}@{\hskip 0.15cm}
    p{5.5 cm}@{\hskip 0.15cm} 
    p{5.5 cm}@{\hskip 0.15cm}
}
\hline
{} & {} & {} & \textbf{AlexNet} & \textbf{ResNet18} \\
\hline
{} & \multicolumn{4}{c}{\textit{\textbf{Number of activations}}}\\
{} & {} & {} & {
\hbox{
$\underbrace{20412}_{\text{\tiny features.2}}$
+ $\underbrace{591408}_{\text{\tiny features.5}}$
+ $\underbrace{5218304}_{\text{\tiny features.7}}$
}
\hbox{
+ $\underbrace{4415488}_{\text{\tiny features.9}}$
+ $\underbrace{3636}_{\text{\tiny avgpool}}$
}
\hbox{
= 10249248
}
}
& {
\hbox{
$\underbrace{134848}_{\text{\tiny maxpool}}$ 
+ $\underbrace{2157568}_{\text{\tiny layer1}}$ 
+ $\underbrace{815360}_{\text{\tiny layer2}}$ 
}
\hbox{
+ $\underbrace{1931776}_{\text{\tiny layer3}}$ 
}
\hbox{
= 5039552
}
} \\
\noalign{\vskip -0.3 cm} 
{} & \multicolumn{4}{c}{\textit{\textbf{Samples per activation}}}\\
{} & {\it sitv/si}  & {$(A_1)$} &
{$2^{13} \cdot (2\cdot 8 + 2) = 147456$} & {$2^{13} \cdot (2\cdot 8 + 2) = 147456$} \\
{} & {} & {$(A_2)$} &
{$2^{13} \cdot (2\cdot 10 + 2) = 180224$} & {$2^{13} \cdot (2\cdot 10 + 2) = 180224$} \\
{} & {\it shpv}  & {$(A_1)$} &
{$2^7 \cdot 15 \cdot 2^3 \cdot 2^3= 122880$} & {$2^7 \cdot 15 \cdot 2^3 \cdot 2^3= 122880$} \\
{} & {}  & {$(A_2)$} &
{$2^7 \cdot 20 \cdot 2^3 \cdot 2^3= 163840$} & {$2^7 \cdot 20 \cdot 2^3 \cdot 2^3= 163840$} \\
\hline
\end{tabular}